\documentclass[runningheads]{llncs}
\usepackage{graphicx}
\usepackage{comment}
\usepackage{amsmath,amssymb} % define this before the line numbering.
\usepackage{color}
\usepackage{adjustbox}
\usepackage[width=122mm,left=12mm,paperwidth=146mm,height=193mm,top=12mm,paperheight=217mm]{geometry}

\usepackage[ruled,vlined]{algorithm2e}
\newcommand{\TV}{\texttt{TV}}
\usepackage[software,hardware]{mymacros}
\DeclareMathOperator*{\argminB}{argmin}   % Jan Hlavacek
\usepackage{hyperref}

\usepackage{cleveref}
\Crefformat{figure}{#2Fig.~#1#3}
\Crefmultiformat{figure}{Figs.~#2#1#3}{ and~#2#1#3}{, #2#1#3}{ and~#2#1#3}

\begin{document}
% \renewcommand\thelinenumber{\color[rgb]{0.2,0.5,0.8}\normalfont\sffamily\scriptsize\arabic{linenumber}\color[rgb]{0,0,0}}
% \renewcommand\makeLineNumber {\hss\thelinenumber\ \hspace{6mm} \rlap{\hskip\textwidth\ \hspace{6.5mm}\thelinenumber}}
% \linenumbers
\pagestyle{headings}
\mainmatter
\def\ECCVSubNumber{6631}  % Insert your submission number here

\title{Low-Rank and Total Variation Regularization and Its Application to Image Recovery} % Replace with your title

% INITIAL SUBMISSION 
%\begin{comment}
% \titlerunning{ECCV-20 submission ID \ECCVSubNumber} 
% \authorrunning{ECCV-20 submission ID \ECCVSubNumber} 
% \author{Anonymous ECCV submission}
% \institute{Paper ID \ECCVSubNumber}
%\end{comment}
%******************

% CAMERA READY SUBMISSION
%\begin{comment}
\titlerunning{LR and TV Regularization and Its Application to Image Recovery}
%If the paper title is too long for the running head, you can set
%an abbreviated paper title here

\author{Pawan Goyal\inst{1} \and
Hussam Al Daas\inst{1} \and
Peter Benner\inst{1}}

\authorrunning{P. Goyal et al.}
%First names are abbreviated in the running head.
%If there are more than two authors, 'et al.' is used.

\institute{Max Planck Institute for Dynamics of Complex Technical Systems, \\ Sandtorstr. 1, 39106 Magdeburg, Germany\\
\email{\{goyalp,aldaas,benner\}@mpi-magdeburg.mpg.de}}
%\end{comment}
%******************
\maketitle

\begin{abstract}
In this paper, we study the problem of image recovery from given partial (corrupted) observations. Recovering an image using a low-rank model has been an active research area in data analysis and machine learning. But often, images are not only of low-rank but they also exhibit sparsity in a transformed space.  In this work, we propose a new problem formulation in such a way that we seek to recover an image that is of low-rank and has sparsity in a transformed domain. We further discuss various non-convex non-smooth surrogates of the rank function, leading to a relaxed problem. Then, we present an efficient iterative scheme to solve the relaxed problem that essentially employs the (weighted) singular value thresholding at each iteration. Furthermore, we discuss the convergence properties of the proposed iterative method. We perform extensive experiments, showing that the proposed algorithm outperforms state-of-the-art methodologies in recovering images.
\end{abstract}
\section{Introduction}
Low-rank matrix recovery from partial  (corrupted) observations has been intensively studied due to its vast applications in computer vision and machine learning. For instance, in a recommender system, the data matrix exhibits low-rank properties since a few factors play a role in the preferences of a customer, see e.g., \cite{srebro2010collaborative}; human facial images can be approximated very well by a low-dimensional linear subspace, therefore, a corrupted facial image can be recovered under the hypothesis that all the images lie in a low-dimensional subspace. Moreover, consider that a video is taken with a static background and has a small moving part such as a car or a person. Then, one may ask if it is possible to extract the background (as a low-rank term) and foreground (as a sparse term) information, see, e.g., \cite{mu2011accelerated,wright2009robust}. 

Algorithms that recover an underlying low-rank structure can be broadly characterized  in two categories. In one category, we assume an explicit low-rank form of the solution $\bX$, meaning that $\bX \in \Rnm$ can be decomposed as a product of two smaller matrices $\bX_1\in \Rnr, \bX_2\in \Rmr$, i.e., $\bX \approx \bX_1\bX_2^T$, see \cite{buchanan2005damped,eriksson2010efficient,ke2005robust,srebro2003weighted}. One drawback of these algorithms is that they need a prior estimate of the rank of the solution which is hard to be estimated in advance. In the other category, the problem is defined directly using the rank function of the solution $\bX$. Nevertheless, optimization problems involving the rank function are known to be NP-hard. Hence, they are not practical when it comes to even medium-sized problems. Therefore, there has been extensive research in replacing the rank function by some surrogate functions. One very popular surrogate is the nuclear-norm of the matrix $\bX$, denoted by $\|\cdot\|_*$, which is defined as the sum of its singular values, i.e., $\|\bX\|_* = \sum_i\sigma_i(\bX)$, where $\sigma_i(\bX)$ are the singular values of the matrix $\bX$. It is shown in \cite{recht2010guaranteed} that  the nuclear-norm is the best convex envelop to the rank function. Nuclear-norm based surrogate modeling of the rank function  has received a lot of attention due to various reasons. One important reason among others is that there exists a closed-form solution to the following optimization problem:
\begin{equation}\label{eq:NN_prob}
 \min_\bX \lambda \|\bX\|_* + \dfrac{1}{2} \|\bX - \bY\|_F^2
\end{equation}
that is given by a soft-thresholding operation on the singular values of the matrix $\bY$, i.e., 
\begin{equation}
 \bX^* = \bU\cD_\lambda(\mathbf{\Sigma})\bV^\top,
\end{equation}
where $\bY = \bU\mathbf{\Sigma}\bV^\top$ is the singular value decomposition (SVD) of the matrix $\bY$ with $\mathbf{\Sigma} = \diag{\sigma_1,\ldots,\sigma_n}, \sigma_{i}\geq \sigma_{i+1}$, and \begin{equation}\cD_\lambda(\mathbf{\Sigma}) = \diag{(\sigma_1-\lambda)_+,\ldots,(\sigma_n-\lambda)_+}\end{equation} with $t_+ := \max{(t,0)}$, see \cite{cai2010singular}. It has been proven in \cite{candes2009exact,recht2010guaranteed} that under certain conditions, a low-rank matrix can be recovered using partial or corrupted observations by solving \eqref{eq:NN_prob}. However, when these conditions are not fulfilled, the problem \eqref{eq:NN_prob} might not recover exactly the low-rank solution. To overcome this shortcoming, there have been several attempts, i.e., tighter  surrogates of the rank function have been proposed, see, e.g., \cite{frank1993statistical,friedman2012fast,gao2011feasible,geman1995nonlinear,lu2015nonconvex,trzasko2008highly,zhang2010nearly,zhang2010analysis} and weighted nuclear-norm concepts are proposed in \cite{gu2017weighted,gu2014weighted,srebro2003weighted}.

\begin{figure}[tb]
\centering
\includegraphics[width = 0.4\textwidth]{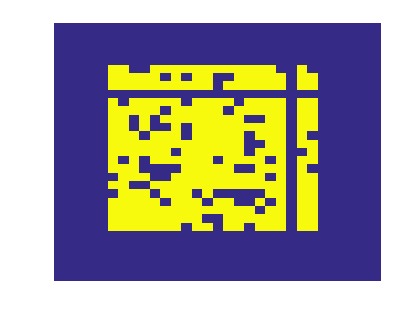}
\caption{An image whose row/columns are completely missing. Recovery of those row/columns is not possibly recoverable using solely a rank-based optimization problem}
\label{fig:missingvalues1}
\end{figure}

\begin{figure}[tb]
\centering
\includegraphics[width = 1.1\textwidth,trim = 2cm 0cm 0cm 0cm,clip]{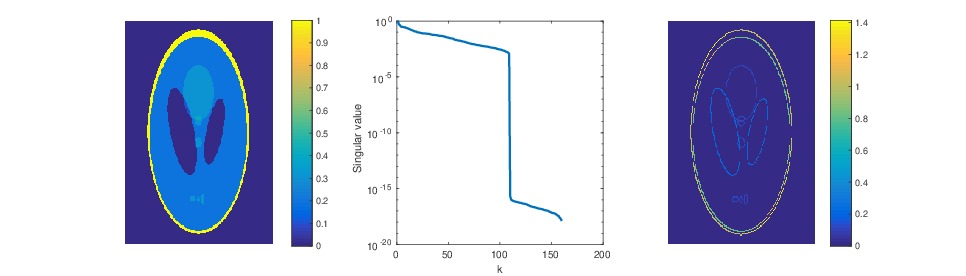}
\caption{Shepp-Logan Phantom: The true image is shown in the left,  the decay of the singular values is shown in the middle, and in the right, we plot the image $\bM_\bX$, defined in~\eqref{eq:mtx_def_tv}}
\label{fig:missingvalues2}
\end{figure}

Although image recovery has been intensively studied using the rank function or its surrogate regularization and has been successful, the problem formulation shares two main issues.
\begin{itemize}
\item First, it will fail to recover an image when a row/column is completely missing as shown for example in \Cref{fig:missingvalues1}.
\item Second, most of the images in practice do not only exhibit low-rank properties. They also exhibit a sparsity property in a transformed space or piece-wise smoothness. 
\end{itemize}
  For piece-wise smoothness, we consider anisotropic total variation, defined as follows:
\begin{equation}\label{eq:TV_def}
 \|\bX\|_\TV := \sum_{i,j} \bM_\bX(i,j)) =:  \|\bM_\bX\|_{l_{1,1}},
\end{equation}
where \begin{equation} \label{eq:mtx_def_tv}
\bM_\bX(i,j) := |\bX(i,j) - \bX(i,j+1)| +   |\bX(i+1,j) - \bX(i,j)|       
      \end{equation}
and $\bX(i,j)$ denotes the $(i,j)$th entry of the matrix $\bX$. To illustrate the sparsity and low-rank phenomenon, we consider an image, known as \emph{Shepp-Logan Phantom}, related to the medical applications. In \Cref{fig:missingvalues2}, we plot the image, the decay of the singular values that indicates whether the image is of low-rank, and the entries of the $\bM_\bX$ indicates the sparsity of a transformed space. The figure shows that the image does not have a fully low-rank characteristic although the singular values decay rapidly, and the image is rather sparse in the transformed space that defines the total variation. Therefore, we can expect a better image recovery if a recovery problem is regularized using a combination of the rank and total variation functions. 

In this paper, we study the recovery of images under partial or corrupted observations. Towards this, we propose an optimization problem using a regularizer that is a combination of a surrogate function of the  rank function and total variation. However, solving the proposed optimization problem is a big challenge because of its non-convex non-smoothness nature. So, we also discuss an efficient iterative scheme to solve the problem that is essentially based on singular value thresholding and its variant.

The rest of the paper is structured as follows. In \Cref{sec:problemformulation}, we formulate an optimization problem for image recovery. We further propose an iterative scheme to solve the optimization problem efficiently and discuss its convergence. In \Cref{sec:numerics}, we present experimental studies and show that the proposed method outperforms state-of-the-art algorithms in both peak signal-to-noise ratio (PSNR) and preserving local features. We conclude the paper in \Cref{sec:conclusions}. 
 
\section{Low-Rank and Total Variation Regularized Problem}\label{sec:problemformulation}
\subsection{Problem Formulation}
In this section, we discuss a problem formulation for image recovering from  partial or corrupted observations. Using an appropriate prior hypothesis about an image, we can expect to have a better recovery. Towards this, we seek to regularize a recovery problem in such a way that allows us to reconstruct local information  of an image  (captured by the total variation) as well as global information (captured by the rank-based regularization). For this reason, we propose the following regularized problem: 
 \begin{equation}\label{eq:OptProb_Rank}
   \min_\bX     \lambda_1 \rank{\bX} + \lambda_2 \|\bX\|_\TV  + \| \bP_\ensuremath{\Omega}(\bX-\bM)\|_F,
 \end{equation}
where $\Omega$ is a set of observed indices, $\bP_\Omega$ is an orthonormal projector such that $\bP_\Omega(\bX) = \bX(i,j)$ if $(i,j) \in \Omega$ and zero otherwise, $\|\cdot\|_\TV$ is defined in \eqref{eq:TV_def} which encodes spatially local information of the image $\bX$, and $\rank{\bX}$ gives us a global information about the image. Having the first two terms in the optimization problem \eqref{eq:OptProb_Rank} aims at taking into account both local and global information; the parameters $\lambda_{\{1,2\}}$ define the weighting to these information. 

\begin{figure}[tb]
\centering
 \includegraphics[scale = 0.2]{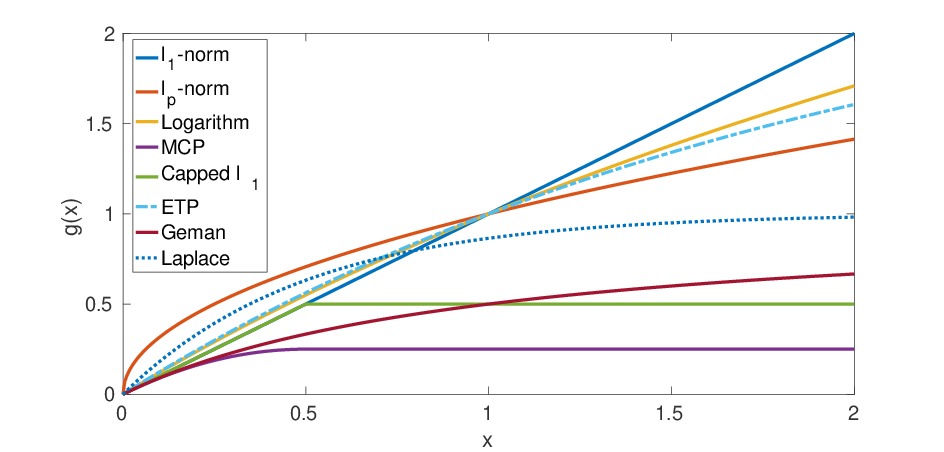}
 \caption{A pictorial perspective of the surrogate functions listed in \cref{tab:surrogatefunctions}  for $\lambda~=~1, \gamma~=~0.5$}
 \label{fig:pic_surrogate}
\end{figure}

In general, optimization problems involving the rank function are known to be combinatorial NP hard.
As a remedy, we seek to solve a relaxed problem that is obtained by replacing the rank constraint by an appropriate surrogate function.
Notice  that the rank function of a matrix $\bX$ is the $l_0$-norm of the vector of the singular values of the matrix $\bX$, i.e., $\rank{\bX} = \|\sigma\|_{l_0}$, where $\sigma = \begin{bmatrix} \sigma_1,\ldots,\sigma_n\end{bmatrix}$ in which the $\sigma_i$'s are the singular values of the matrix $\bX$ sorted by magnitude, $\sigma_{i} \geq \sigma_{i+1}$. 
  Inspired from compressed sensing \cite{candes2006robust,donoho2006compressed}, the $l_1$-norm of the singular values, i.e., $\sum_i\sigma_i =:\|\bX\|_*$ can be a suitable surrogate of the $l_0$-norm. An appealing feature of $l_1$-norm or the nuclear-norm minimization is that the relaxed optimization problem becomes convex which can be solved very efficiently. Despite a success of the $l_1$-relaxation in recovering $l_0$ solutions, it is known that the $l_1$-norm is a loose approximation to the $l_0$-norm. Recently, non-convex non-smooth surrogates to the $l_0$-norm have received much attention. Some of the popular surrogate functions of $l_0$-norm are listed in \Cref{tab:surrogatefunctions} and in \Cref{fig:pic_surrogate}, we provide a pictorial perspective of these surrogate functions. 

\begin{table}[tb]
\centering
\caption{Surrogate approximation functions of $\|x\|_{l_0}$ for $x \geq 0$ and $\lambda,\gamma > 0$}
\label{tab:surrogatefunctions}
 \begin{tabular}{|rl|}  \hline
 $L_1$-norm: & $\lambda x$ \\ \hline
  $L_p$-norm \cite{frank1993statistical}: & $\lambda x^p$ \\ \hline
Logarithm \cite{friedman2012fast}: & $\tfrac{\lambda}{\log{\gamma + 1}} \log{\gamma x+1}$  \\ \hline
Minimax concave penalty (MCP) \cite{zhang2010nearly}: &  $\begin{cases}
\lambda x - \tfrac{x^2}{2\gamma} & \text{if}~ x < \lambda\gamma \\
\tfrac{1}{2}\gamma\lambda^2 &\text{if}~ x\geq \gamma\lambda
\end{cases}
$ \\ \hline
Capped $l_1$ \cite{zhang2010analysis}: & $\begin{cases} \lambda x & x < \gamma \\  \lambda \gamma &  x \geq \gamma
 \end{cases} $ 
 \\ \hline 
 Exponential type penalty (ETP) \cite{gao2011feasible}: & $\dfrac{\lambda}{1-\exp(-\gamma)} \left(1-\exp(-\gamma x) \right)$ 
 \\ \hline
 Geman \cite{geman1995nonlinear}: & $\tfrac{\lambda x}{x + \gamma}$ \\ \hline
 Laplace \cite{trzasko2008highly}:& $\lambda \left(1-\exp\left(-\tfrac{x}{\gamma}\right) \right)$ \\  \hline
 \end{tabular}
\end{table}

Consequently, we seek to solve a relaxation of the problem \eqref{eq:OptProb_Rank} by replacing the rank function  by a surrogate function using the singular values of the solutions. Precisely, we aim at solving 
\begin{equation} \argminB_\bX\cE(\bX), \end{equation}
where 
 \begin{equation}\label{eq:OptProb_Surrogate}
  \cE(\bX) :=  \left(  \lambda_1 \sum_{i}g(\sigma_i) + \lambda_2 \|\bX\|_\TV  + \| \bP_\ensuremath{\Omega}(\bX) - \bP_\Omega(\bM)\|_F\right),
 \end{equation}
and $g(\cdot)$ is a non-convex non-smooth surrogate function of the $l_0$-norm.  In the subsequent subsection, we discuss an iterative scheme that aims at solving \eqref{eq:OptProb_Surrogate}.

\subsection{Optimization Scheme}
We first assume that the function $g(\cdot)$ is concave and a monotonically increasing function. Thus, we have
\begin{equation}\label{eq:concavepro}
 g(z)  \leq g(z^{k}) + \langle s^k, z - z^k\rangle,
\end{equation}
where $s^k \in \partial g(z^k)$ with $\partial g(z^k)$ denoting its super-gradient at $z_k$, see, e.g., \cite{rockafellar1970convex}. Using the property \eqref{eq:concavepro}, we can arrive at a subproblem that generates the sequence of $\bX^{k}$, leading us to the optimal solution if it converges. That is,
\begin{equation}\label{eq:opt_solution}
\begin{aligned}
\bX^{k+1} :=  \argminB_{\bX}\left(\sum_{i=1}^n \lambda_1\left(g(\sigma_i^k) + \langle s_i^k, \sigma_i - \sigma_i^k\rangle \right) + \lambda_2 \|\bX\|_\TV\right) + \| \bP_\ensuremath{\Omega}(\bX - \bM) \|_F,
\end{aligned}
\end{equation}
where $s_i^k$ is the super-gradient of the function $g(\cdot)$ at $\sigma^k_i$ and the $\sigma_i^k$'s are the singular values of $\bX^k$ --- the solution of the subproblem at the previous step --- and $\sigma_i$ denotes the singular values of $\bX$. Since $g(\sigma_i^k)$ and $s_i^k\sigma_i^k$ are constants, \eqref{eq:opt_solution} boils down to 
\begin{equation}\label{eq:opt_solution1}
\begin{aligned}
\bX^{k+1} :=  \argminB_{\bX}\left(\sum_{i=1}^n w^k_i\sigma_i + \lambda_2 \|\bX\|_\TV\right) + \| \bP_\ensuremath{\Omega}(\bX - \bM) \|_F,
\end{aligned}
\end{equation}
where $w^k_i := \lambda_1s_i^k$. Note that the problem \eqref{eq:opt_solution1} is still non-convex and not easy to solve. To ease the problem further to solve for $\bX^{k+1}$, we linearize the last two terms of \eqref{eq:opt_solution1} and add a proximal term. This yields
\begin{align}
  \cF(\bX) := &\lambda_2 \|\bX\|_\TV + \| \bP_\ensuremath{\Omega}(\bX - \bM) \|_F \nonumber\\
  &\qquad \approx \lambda_2 \|\bX^k\|_\TV + \| \bP_\ensuremath{\Omega}(\bX^k - \bM) \|_F + \left\langle t^k, \bX-\bX^k \right\rangle + \dfrac{\mu}{2}\|\bX-\bX^k\|_F^2, \label{eq:linearize}
\end{align}
where $t^k$ is the sub-gradient of the function ${\cF}(\bX)$ at $\bX^k$, and $\mu > 0$ is the proximal parameter. As a result, for the update $\bX^{k+1}$, we solve the following optimization problem:
\begin{align}
 \bX^{k+1} &= \min_{\bX}  \left(\sum_{i=1}^n w_i\sigma^{k+1}_i + \left\langle t^k, \bX^{k+1}-\bX^k  \right\rangle + \dfrac{\mu}{2} \|\bX^{}-\bX^k \|_2 \right) \nonumber\\
  &= \min_{\bX^{}}  \left(\sum_{i=1}^n  w_i\sigma^{k+1}_i +  \dfrac{\mu}{2} \left\|\bX^{}- \left(\bX^k - \dfrac{1}{\mu} t^k\right)  \right\|_2 \right).\label{eq:Prob_WNN}
 \end{align}
 Note that $w^k_i \leq w^k_{i+1}$, or $s^k_{i} \leq s^k_{i+1}$ due to the concavity assumption on the surrogate function  $g(\cdot)$. Interestingly,  there exists an analytic solution to the optimization problem \eqref{eq:Prob_WNN}  although the problem is still non-convex. The solution of the optimization problem can be given by the singular value thresholding. In the following, we recall the result from \cite{chen2013reduced}. 

\begin{theorem}
 Consider $\lambda > 0$ and a matrix $\bY \in \Rnm$. Moreover, let us assume that $0 \leq w_1\leq \cdots \leq w_n$ Then, a globally optimal solution to the following problem 
 \begin{equation}
  \min  \sum_{i=1}^n w_i\sigma_i + \dfrac{1}{2}\|\bX - \bY\|_F^2
 \end{equation}
is given by the weighted singular value thresholding 
\begin{equation}\label{eq:shrinkage}
 \bX^* = \bU\bS_{w}(\mathbf{\Sigma})\bV^T,
\end{equation}
where $\bY = \bU\mathbf{\Sigma} \bV^T$ is the SVD of $\bY$ and \begin{equation}\bS_{ w}(\mathbf{\Sigma}) = \diag{\left(\sigma_1 -  w_1\right)_+,\ldots,\left(\sigma_n -  w_n\right)_+} \end{equation} with $t_+ := \max(t,0)$.
\end{theorem}

Finally, we summarize all necessary steps in \Cref{alg:LRTV_IterativeAlgo} that generates the sequence $\bX^{k+1}$ and gives an optimal solution to the problem \eqref{eq:OptProb_Surrogate} if it convergences. 
\begin{algorithm}[tb]
\SetAlgoLined
\KwIn{Initial guess $\bX_0$, $\mu$, $k=0$.}
 \While{Until convergence}{
 Compute the sub-gradient of the function $g(\cdot)$, i.e., $w_i^k = \partial g(\sigma_i^k)$. \\
 Compute $t^k$ as defined in \eqref{eq:linearize}.\\ 
 Define $\bY \leftarrow \bX^{k} - \dfrac{1}{\mu} t^k.$\\
 Compute $\bX^{k+1} \leftarrow \bU\cS_{w}(\mathbf{\Sigma})\bV^T$, where $\bY = \bU\mathbf{\Sigma}\bV^T$ denotes the SVD of $\bY$, and $\cS_{w}(\cdot)$ is the shrinkage operator defined in \eqref{eq:shrinkage}. \\
 $k \leftarrow k+1.$
 }
     \caption{An iterative procedure to solving image completion using a low-rank and total variation regularization. }
     \label{alg:LRTV_IterativeAlgo}
\end{algorithm}

\subsection{Some Remarks}
\begin{remark}
 Note that the gradients of the functions $f_1(\bX) := \lambda_2\|\bX\|_\TV$ and $f_2(\bX) := \|\bP_\Omega(\bX-\bM)\|_F$ are Lipschitz continuous. Thus, we can write
\begin{equation}
 \|\partial f(\bX) -  \partial f(\bY) \|_F \leq \beta \|\bX-\bY\|_F, \quad \forall \bX, \bY \in \Rmn.
\end{equation}
where $f(\bX) := f_1(\bX) +f_2(\bX)$,  $\partial$ denotes the sub-gradient operator, and $\beta$ is the Lipschitz constant of $\partial f(\bX)$. If the proximity parameter $\mu$ in \eqref{eq:linearize} is greater than $\beta$, then we have $\lim_{k \rightarrow \infty} \|\bX^{k+1} - \bX^k\|_F = 0$. This result directly follows from~\cite{lu2015nonconvex}. In fact, it can also be proven that the objective function defined in~\eqref{eq:OptProb_Surrogate} is a non-increasing function, i.e., $\cE(X^{k+1}) \leq \cE(X^k)$.
\end{remark}

\begin{remark}\label{rem:rankdec}
  We note an interesting point: if we choose the function $g(x)$ such that $\partial g(0) = \infty$, e.g., $g(x) = x^p, 0<p<1$, then the sequence generated by \Cref{alg:LRTV_IterativeAlgo} will be of non-increasing rank as well. From the previous remark, it also follows that the objective function is also a non-increasing function. 
\end{remark}

\begin{remark}
 To solve the following problem:
 \begin{equation}\label{eq:OptProb_Rank1}
   \min_\bX \lambda_1 \rank{\bX} + \lambda_2 \|\bX\|_\TV \quad \text{subject to}~~   \bP_\ensuremath{\Omega}(\bX - \bM) = 0,% {\color{red}\leq \varepsilon},
 \end{equation}
 we propose an iterative scheme as shown in \Cref{alg:LRTV_IterativeAlgo1}. Note that the optimal solution at each step is considered as the initial guess at the next step. Convergence analysis of \Cref{alg:LRTV_IterativeAlgo1} is much more involved and is beyond the scope of this paper. 
 \begin{algorithm}[tb]
 \SetAlgoLined
  initialization $0<\alpha <1, \texttt{tol},k = 0$\;
  \While{$\alpha^k < \texttt{tol}$}{
  Solve
    $\argminB_\bX  \alpha^k \left(\lambda_{1}g(\sigma_i) + \lambda_2 \|\bX\|_\TV\right) + \| \bP_{\mathbf{\Omega}}(\bX) -\bM) \|_F$ using \Cref{alg:LRTV_IterativeAlgo}.\\
   $k \leftarrow k +1$
  }
    \caption{An iterative procedure to solve the problem \eqref{eq:OptProb_Rank1}}
     \label{alg:LRTV_IterativeAlgo1}
 \end{algorithm}

 \end{remark}

\section{Numerical Experiments}\label{sec:numerics}
In this section, we present numerical experiments to assess the effectiveness of our proposed algorithm (denoted by IRNN\_TV) and compare it to other state-of-the-art methods.
The data set in our experiments contains synthetic data arising from medical imaging and academic test cases.
All experiments are performed using \matlab~2019b.
We compare our method against three other methods.
The first method is based only on the total variation minimization. This method is proposed in~\cite{TFOCS} and available through the {TFOCS} package.
The second is based only on a low-rank minimization technique. It is presented in~\cite{lu2015nonconvex} and available through the {IRNN} package.
The last method is called LMaFit~\cite{LMaFit}. It aims at fitting a low-rank matrix such that it approximates the known entries of the matrix needed to be recovered.

These methods are compared based on their effectiveness in recovering images obtained from a set of test images after modifying them by either removing some entries (a fraction of its size) or removing some entries and adding noise to the rest. 
These two problems correspond to image completion without and with noise in observations.

\subsection{Parameter Set-Up}
Here, we explain the parameter set-up of the algorithms that we use in the numerical experiments.
Due to the limit of space, we present numerical experiments by using only one surrogate function ($l_p$ with $p = 0.5$~\Cref{tab:surrogatefunctions}) which gave the best results for both IRNN and our method.
In~\Cref{alg:LRTV_IterativeAlgo,alg:LRTV_IterativeAlgo1}, we set $\lambda_1 =\|\bP_\Omega(\bM)\|_F$, $\lambda_2 = 0.02 \|\bP_\Omega(\bM)\|_F$, and $\alpha = 0.9$. 
Concerning LMaFit, we set the maximal number of iterations to $10,000$ and use the rank increasing strategy with an estimated rank of $50$.
Maximal iteration count of 1,000 is set for TFOCS.
The stopping criterion is either reaching the maximal iteration number or reaching a residual norm less than a predefined threshold ($10^{-6}$ for  observations without noise and the Frobenius-norm of the noise for   observations with noise).
\subsection{Image Completion}
In this section, we consider the recovery of an image starting from partial data which are observed exactly.
We vary the fraction of observed data in the set $\{ 0.2, 0.5\}$ and compare our proposed method against the methods mentioned previously which were introduced to tackle such a problem.

\begin{figure}[tb]
  \begin{minipage}{0.16\linewidth}
    \begin{adjustbox}{width=\linewidth}
        \includegraphics{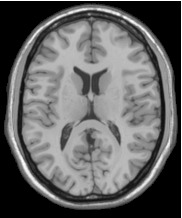}
    \end{adjustbox}
  \end{minipage}
  \begin{minipage}{0.16\linewidth}
    \begin{adjustbox}{width=\linewidth}
        \includegraphics{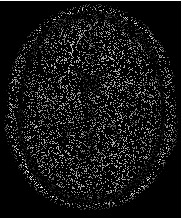}
    \end{adjustbox}
  \end{minipage}
  \begin{minipage}{0.16\linewidth}
    \begin{adjustbox}{width=\linewidth}
        \includegraphics{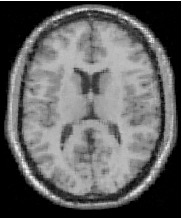}
    \end{adjustbox}
  \end{minipage}
  \begin{minipage}{0.16\linewidth}
    \begin{adjustbox}{width=\linewidth}
        \includegraphics{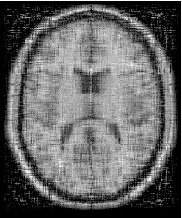}
    \end{adjustbox}
  \end{minipage}
  \begin{minipage}{0.16\linewidth}
    \begin{adjustbox}{width=\linewidth}
        \includegraphics{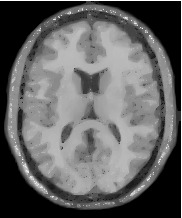}
    \end{adjustbox}
  \end{minipage}
  \begin{minipage}{0.16\linewidth}
    \begin{adjustbox}{width=\linewidth}
        \includegraphics{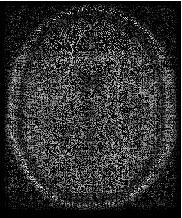}
    \end{adjustbox}
  \end{minipage}

  \begin{minipage}{0.16\linewidth}
    \begin{adjustbox}{width=\linewidth}
        \includegraphics{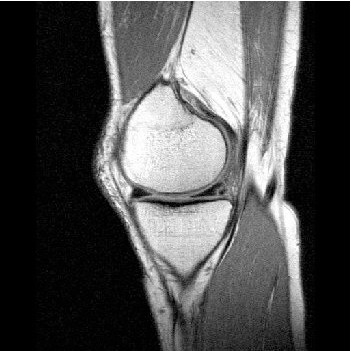}
    \end{adjustbox}
  \end{minipage}
  \begin{minipage}{0.16\linewidth}
    \begin{adjustbox}{width=\linewidth}
        \includegraphics{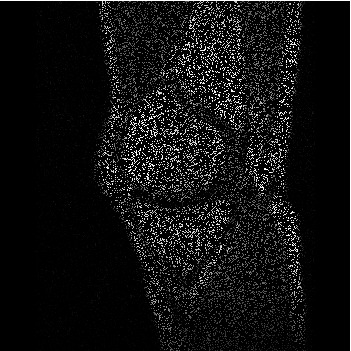}
    \end{adjustbox}
  \end{minipage}
  \begin{minipage}{0.16\linewidth}
    \begin{adjustbox}{width=\linewidth}
        \includegraphics{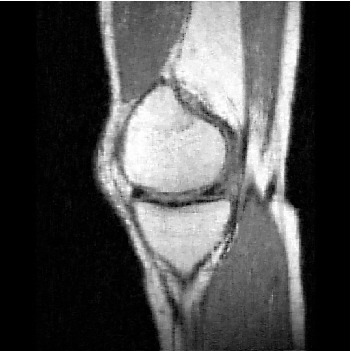}
    \end{adjustbox}
  \end{minipage}
  \begin{minipage}{0.16\linewidth}
    \begin{adjustbox}{width=\linewidth}
        \includegraphics{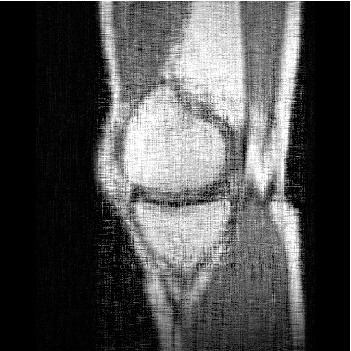}
    \end{adjustbox}
  \end{minipage}
  \begin{minipage}{0.16\linewidth}
    \begin{adjustbox}{width=\linewidth}
        \includegraphics{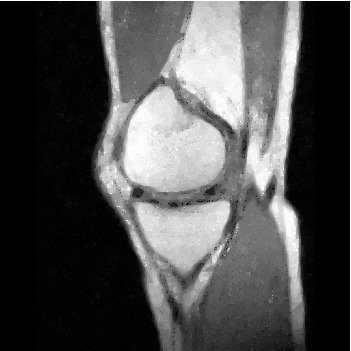}
    \end{adjustbox}
  \end{minipage}
  \begin{minipage}{0.16\linewidth}
    \begin{adjustbox}{width=\linewidth}
        \includegraphics{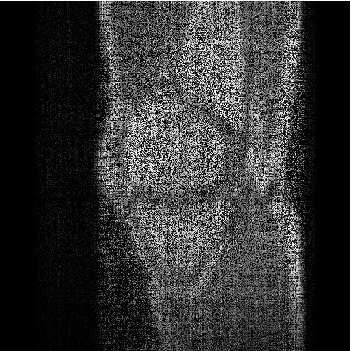}
    \end{adjustbox}
  \end{minipage}

  \begin{minipage}{0.16\linewidth}
    \begin{adjustbox}{width=\linewidth}
        \includegraphics{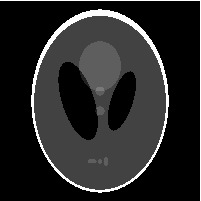}
    \end{adjustbox}
  \end{minipage}
  \begin{minipage}{0.16\linewidth}
    \begin{adjustbox}{width=\linewidth}
        \includegraphics{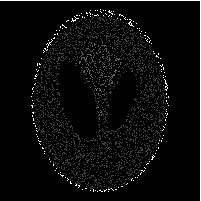}
    \end{adjustbox}
  \end{minipage}
  \begin{minipage}{0.16\linewidth}
    \begin{adjustbox}{width=\linewidth}
        \includegraphics{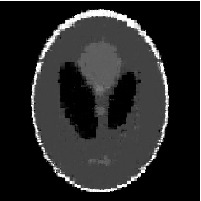}
    \end{adjustbox}
  \end{minipage}
  \begin{minipage}{0.16\linewidth}
    \begin{adjustbox}{width=\linewidth}
        \includegraphics{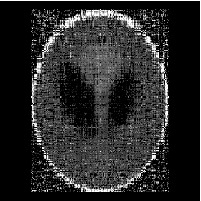}
    \end{adjustbox}
  \end{minipage}
  \begin{minipage}{0.16\linewidth}
    \begin{adjustbox}{width=\linewidth}
        \includegraphics{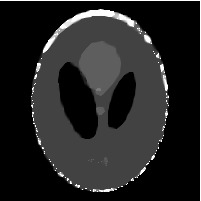}
    \end{adjustbox}
  \end{minipage}
  \begin{minipage}{0.16\linewidth}
    \begin{adjustbox}{width=\linewidth}
        \includegraphics{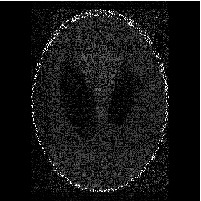}
    \end{adjustbox}
  \end{minipage}

  \begin{minipage}{0.16\linewidth}
    \begin{adjustbox}{width=\linewidth}
        \includegraphics{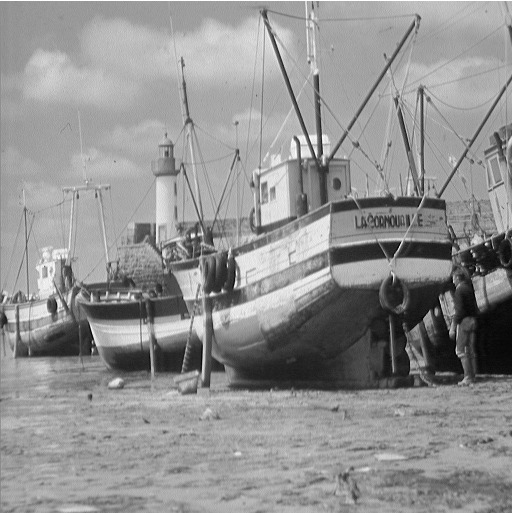}
    \end{adjustbox}
  \end{minipage}
  \begin{minipage}{0.16\linewidth}
    \begin{adjustbox}{width=\linewidth}
        \includegraphics{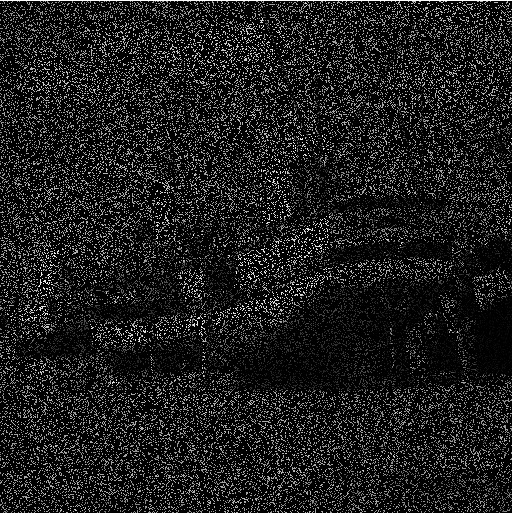}
    \end{adjustbox}
  \end{minipage}
  \begin{minipage}{0.16\linewidth}
    \begin{adjustbox}{width=\linewidth}
        \includegraphics{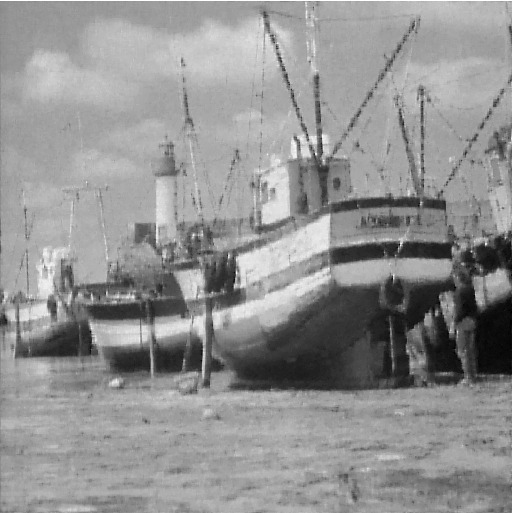}
    \end{adjustbox}
  \end{minipage}
  \begin{minipage}{0.16\linewidth}
    \begin{adjustbox}{width=\linewidth}
        \includegraphics{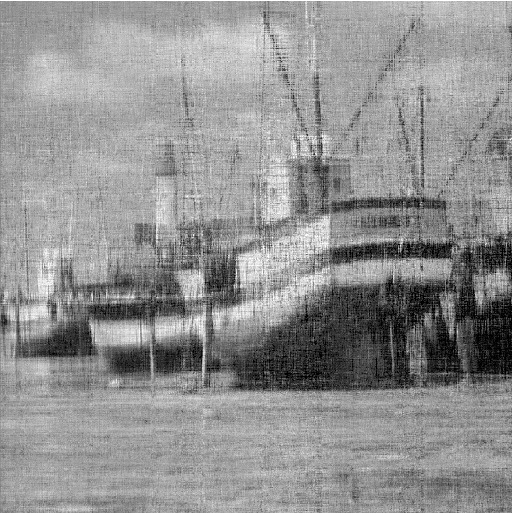}
    \end{adjustbox}
  \end{minipage}
  \begin{minipage}{0.16\linewidth}
    \begin{adjustbox}{width=\linewidth}
        \includegraphics{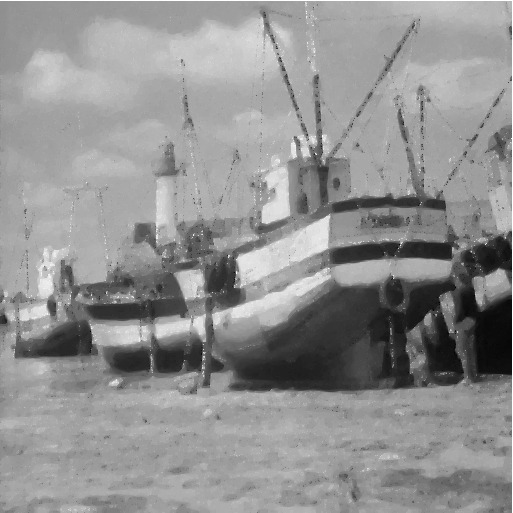}
    \end{adjustbox}
  \end{minipage}
  \begin{minipage}{0.16\linewidth}
    \begin{adjustbox}{width=\linewidth}
        \includegraphics{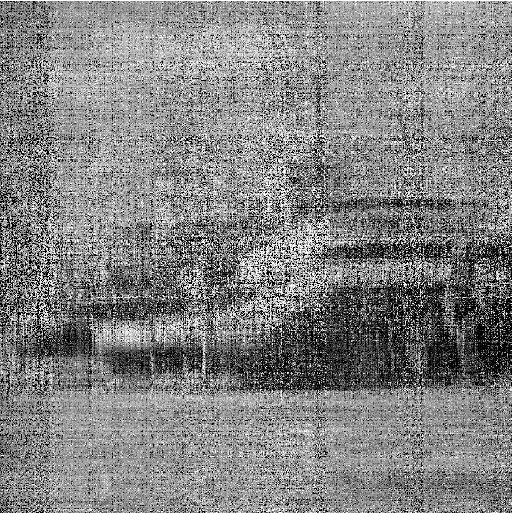}
    \end{adjustbox}
  \end{minipage}

  \begin{minipage}{0.16\linewidth}
    \begin{adjustbox}{width=\linewidth}
        \includegraphics{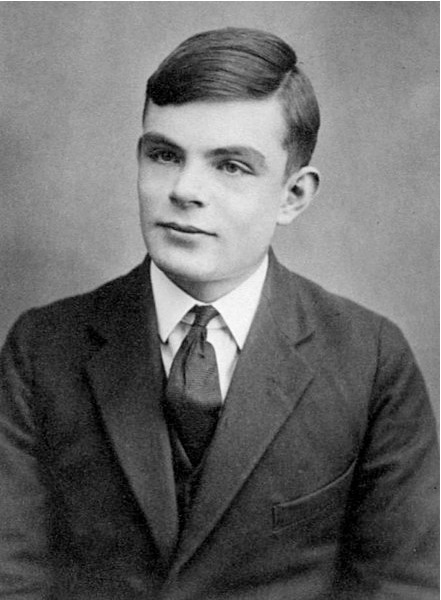}
    \end{adjustbox}
  \end{minipage}
  \begin{minipage}{0.16\linewidth}
    \begin{adjustbox}{width=\linewidth}
        \includegraphics{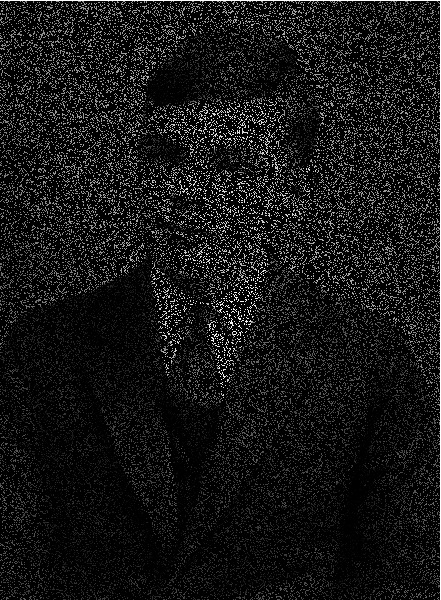}
    \end{adjustbox}
  \end{minipage}
  \begin{minipage}{0.16\linewidth}
    \begin{adjustbox}{width=\linewidth}
        \includegraphics{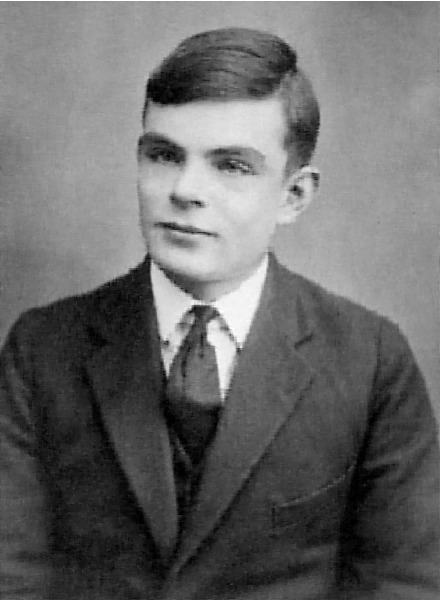}
    \end{adjustbox}
  \end{minipage}
  \begin{minipage}{0.16\linewidth}
    \begin{adjustbox}{width=\linewidth}
        \includegraphics{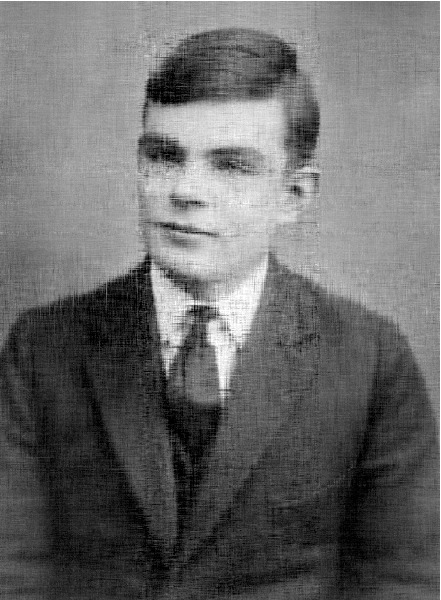}
    \end{adjustbox}
  \end{minipage}
  \begin{minipage}{0.16\linewidth}
    \begin{adjustbox}{width=\linewidth}
        \includegraphics{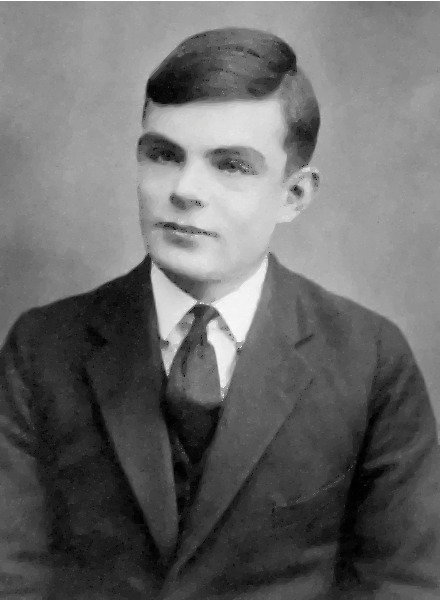}
    \end{adjustbox}
  \end{minipage}
  \begin{minipage}{0.16\linewidth}
    \begin{adjustbox}{width=\linewidth}
        \includegraphics{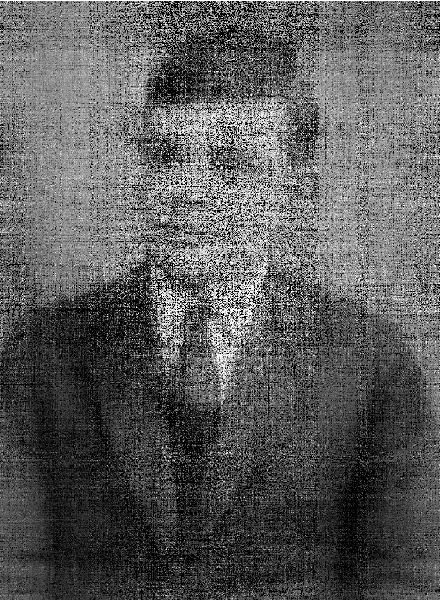}
    \end{adjustbox}
  \end{minipage}
  \caption{Comparison of image recovery by using different techniques when $20\%$ data are observed exactly.
  From the left to the right in each row: original image, observed data, and recovered images by IRNN\_TV, IRNN, TFOCS, LMaFit, respectively}
  \label{fig:completion_20}
\end{figure}

\begin{figure}[tb]
  \begin{minipage}{0.16\linewidth}
    \begin{adjustbox}{width=\linewidth}
        \includegraphics{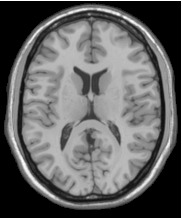}
    \end{adjustbox}
  \end{minipage}
  \begin{minipage}{0.16\linewidth}
    \begin{adjustbox}{width=\linewidth}
        \includegraphics{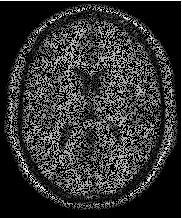}
    \end{adjustbox}
  \end{minipage}
  \begin{minipage}{0.16\linewidth}
    \begin{adjustbox}{width=\linewidth}
        \includegraphics{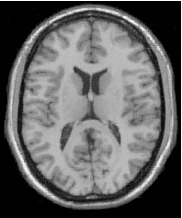}
    \end{adjustbox}
  \end{minipage}
  \begin{minipage}{0.16\linewidth}
    \begin{adjustbox}{width=\linewidth}
        \includegraphics{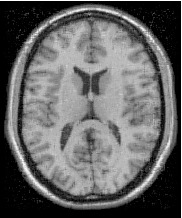}
    \end{adjustbox}
  \end{minipage}
  \begin{minipage}{0.16\linewidth}
    \begin{adjustbox}{width=\linewidth}
        \includegraphics{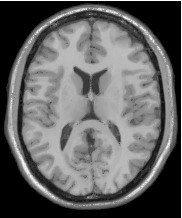}
    \end{adjustbox}
  \end{minipage}
  \begin{minipage}{0.16\linewidth}
    \begin{adjustbox}{width=\linewidth}
        \includegraphics{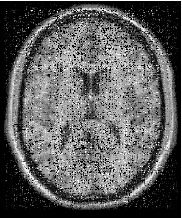}
    \end{adjustbox}
  \end{minipage}

  \begin{minipage}{0.16\linewidth}
    \begin{adjustbox}{width=\linewidth}
        \includegraphics{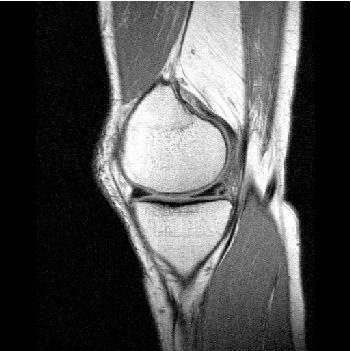}
    \end{adjustbox}
  \end{minipage}
  \begin{minipage}{0.16\linewidth}
    \begin{adjustbox}{width=\linewidth}
        \includegraphics{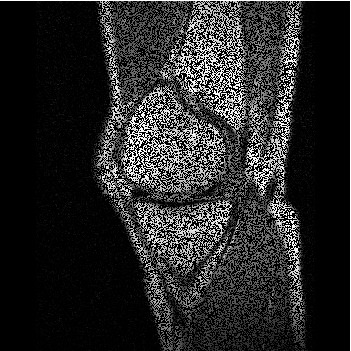}
    \end{adjustbox}
  \end{minipage}
  \begin{minipage}{0.16\linewidth}
    \begin{adjustbox}{width=\linewidth}
        \includegraphics{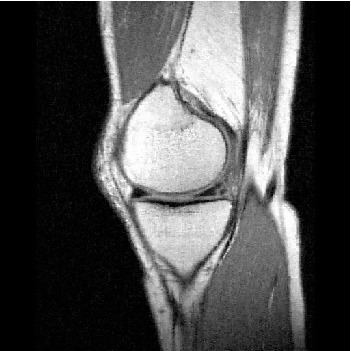}
    \end{adjustbox}
  \end{minipage}
  \begin{minipage}{0.16\linewidth}
    \begin{adjustbox}{width=\linewidth}
        \includegraphics{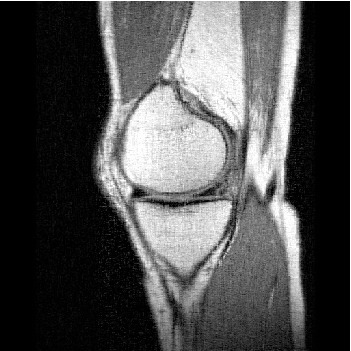}
    \end{adjustbox}
  \end{minipage}
  \begin{minipage}{0.16\linewidth}
    \begin{adjustbox}{width=\linewidth}
        \includegraphics{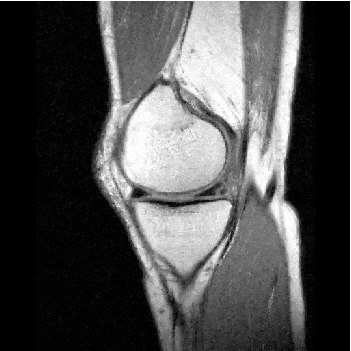}
    \end{adjustbox}
  \end{minipage}
  \begin{minipage}{0.16\linewidth}
    \begin{adjustbox}{width=\linewidth}
        \includegraphics{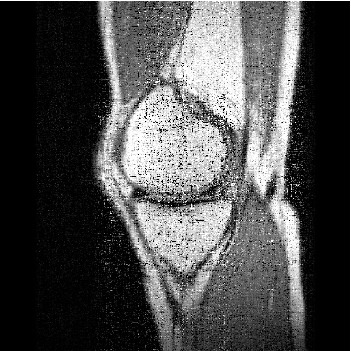}
    \end{adjustbox}
  \end{minipage}

  \begin{minipage}{0.16\linewidth}
    \begin{adjustbox}{width=\linewidth}
        \includegraphics{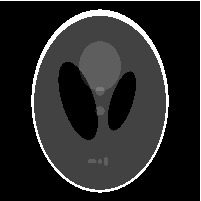}
    \end{adjustbox}
  \end{minipage}
  \begin{minipage}{0.16\linewidth}
    \begin{adjustbox}{width=\linewidth}
        \includegraphics{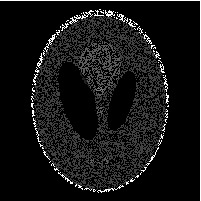}
    \end{adjustbox}
  \end{minipage}
  \begin{minipage}{0.16\linewidth}
    \begin{adjustbox}{width=\linewidth}
        \includegraphics{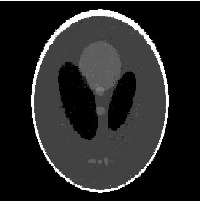}
    \end{adjustbox}
  \end{minipage}
  \begin{minipage}{0.16\linewidth}
    \begin{adjustbox}{width=\linewidth}
        \includegraphics{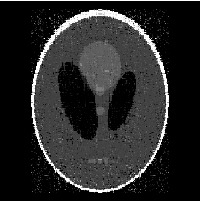}
    \end{adjustbox}
  \end{minipage}
  \begin{minipage}{0.16\linewidth}
    \begin{adjustbox}{width=\linewidth}
        \includegraphics{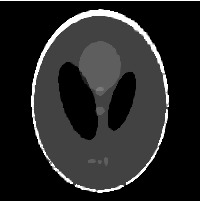}
    \end{adjustbox}
  \end{minipage}
  \begin{minipage}{0.16\linewidth}
    \begin{adjustbox}{width=\linewidth}
        \includegraphics{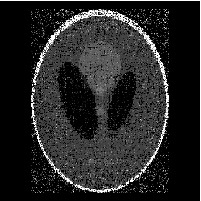}
    \end{adjustbox}
  \end{minipage}

  \begin{minipage}{0.16\linewidth}
    \begin{adjustbox}{width=\linewidth}
        \includegraphics{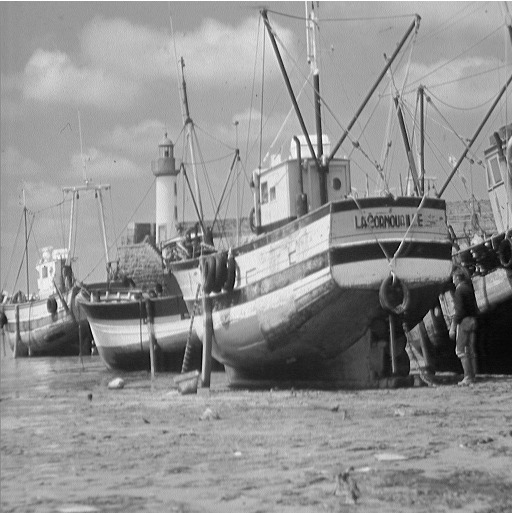}
    \end{adjustbox}
  \end{minipage}
  \begin{minipage}{0.16\linewidth}
    \begin{adjustbox}{width=\linewidth}
        \includegraphics{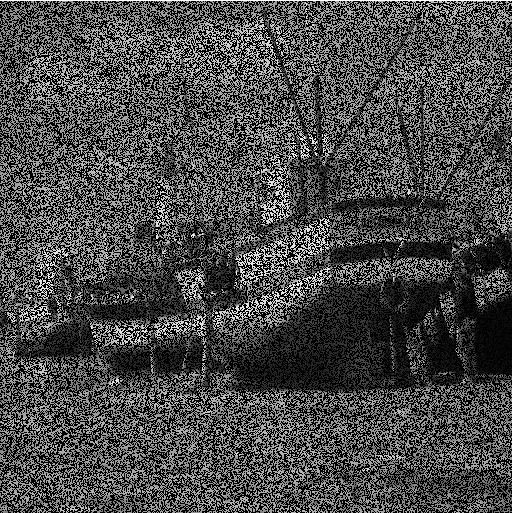}
    \end{adjustbox}
  \end{minipage}
  \begin{minipage}{0.16\linewidth}
    \begin{adjustbox}{width=\linewidth}
        \includegraphics{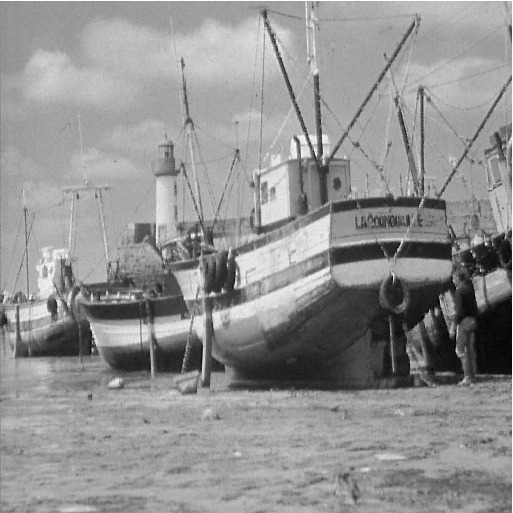}
    \end{adjustbox}
  \end{minipage}
  \begin{minipage}{0.16\linewidth}
    \begin{adjustbox}{width=\linewidth}
        \includegraphics{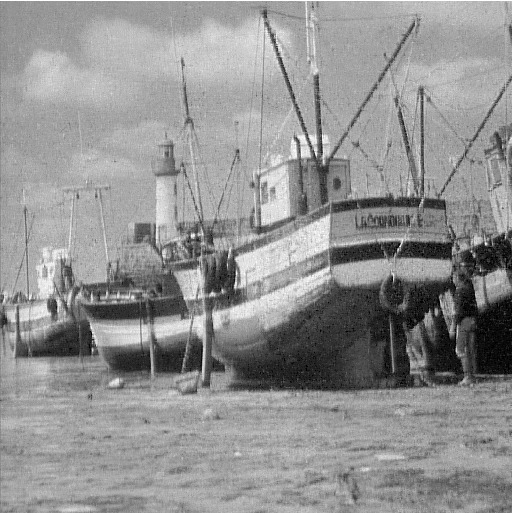}
    \end{adjustbox}
  \end{minipage}
  \begin{minipage}{0.16\linewidth}
    \begin{adjustbox}{width=\linewidth}
        \includegraphics{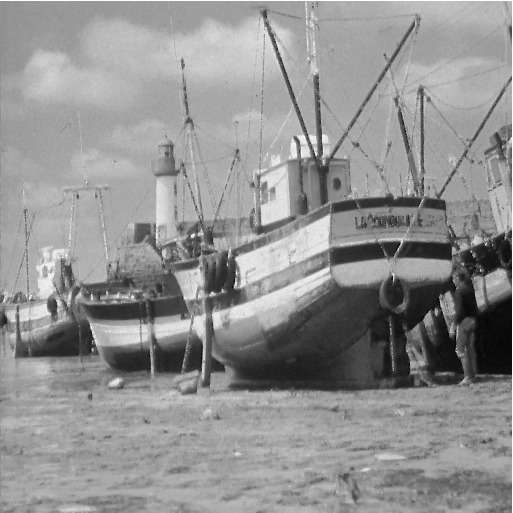}
    \end{adjustbox}
  \end{minipage}
  \begin{minipage}{0.16\linewidth}
    \begin{adjustbox}{width=\linewidth}
        \includegraphics{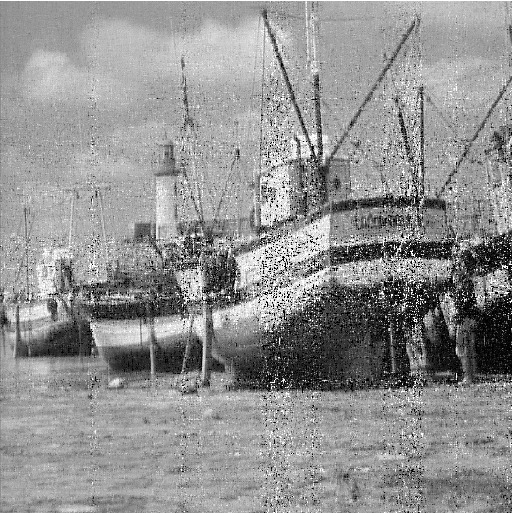}
    \end{adjustbox}
  \end{minipage}

  \begin{minipage}{0.16\linewidth}
    \begin{adjustbox}{width=\linewidth}
        \includegraphics{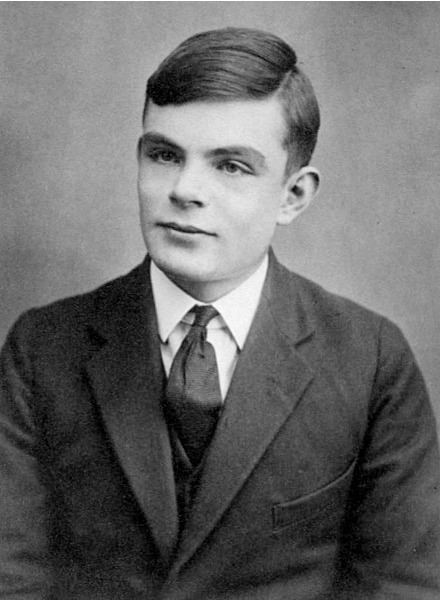}
    \end{adjustbox}
  \end{minipage}
  \begin{minipage}{0.16\linewidth}
    \begin{adjustbox}{width=\linewidth}
        \includegraphics{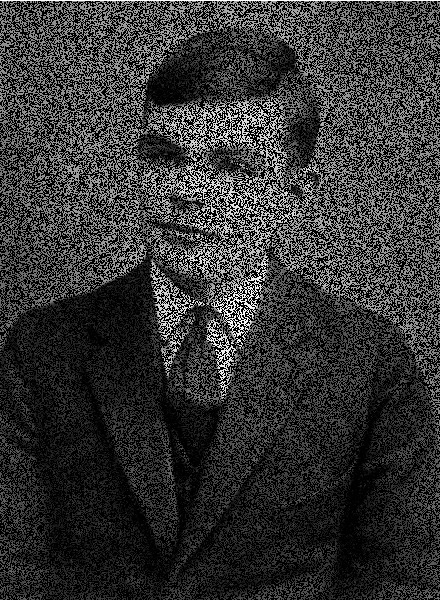}
    \end{adjustbox}
  \end{minipage}
  \begin{minipage}{0.16\linewidth}
    \begin{adjustbox}{width=\linewidth}
        \includegraphics{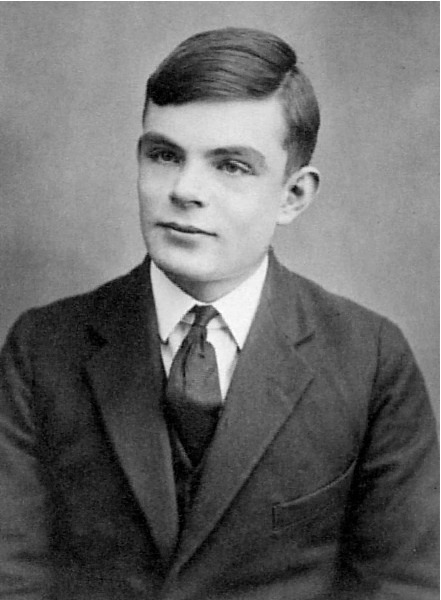}
    \end{adjustbox}
  \end{minipage}
  \begin{minipage}{0.16\linewidth}
    \begin{adjustbox}{width=\linewidth}
        \includegraphics{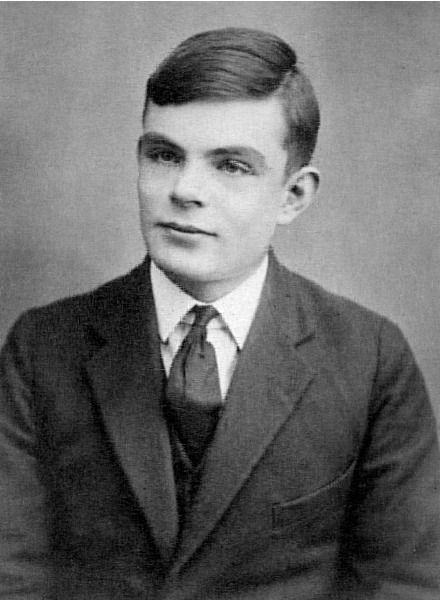}
    \end{adjustbox}
  \end{minipage}
  \begin{minipage}{0.16\linewidth}
    \begin{adjustbox}{width=\linewidth}
        \includegraphics{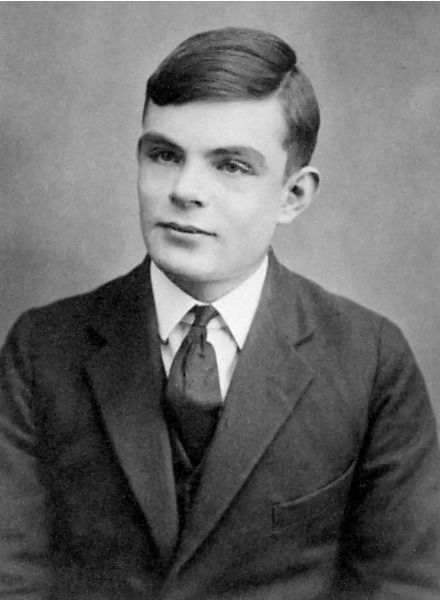}
    \end{adjustbox}
  \end{minipage}
  \begin{minipage}{0.16\linewidth}
    \begin{adjustbox}{width=\linewidth}
        \includegraphics{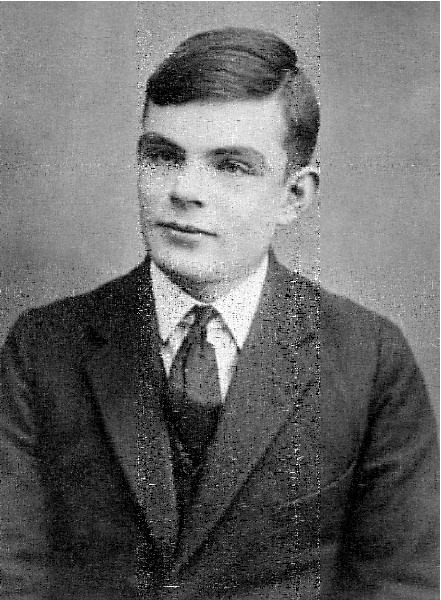}
    \end{adjustbox}
  \end{minipage}
  \caption{Comparison of image recovery by using different techniques when $50\%$ data are observed exactly.
  From the left to the right in each row: original image, observed data, and recovered images by IRNN\_TV, IRNN, TFOCS, LMaFit, respectively}
  \label{fig:completion_50}
\end{figure}

\begin{figure}[tb]
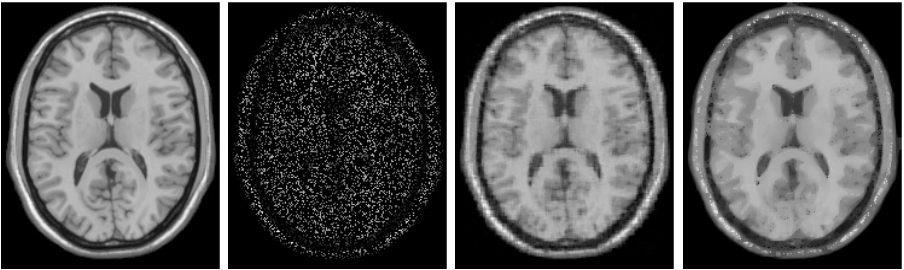

  \begin{minipage}{0.24\linewidth}
    \begin{adjustbox}{width=\linewidth}
        \includegraphics{Figures/completion/20_percent/xch2_1}
    \end{adjustbox}
  \end{minipage}
    \begin{minipage}{0.24\linewidth}
    \begin{adjustbox}{width=\linewidth}
        \includegraphics{Figures/completion/20_percent/xch2_2}
    \end{adjustbox}
  \end{minipage}
  \begin{minipage}{0.24\linewidth}
    \begin{adjustbox}{width=\linewidth}
        \includegraphics{Figures/completion/20_percent/xch2_3}
    \end{adjustbox}
  \end{minipage}
  \begin{minipage}{0.24\linewidth}
    \begin{adjustbox}{width=\linewidth}
        \includegraphics{Figures/completion/20_percent/xch2_5}
    \end{adjustbox}
  \end{minipage}
  \caption{Comparison of image recovery by using our method IRNN-TV and TFOCS when $20\%$ data are observed exactly.
  From the left to the right: original image, observed data, and recovered image by IRNN\_TV and TFOCS, respectively}
  \label{fig:completion_20_percent_TFOCS_IRNN_TV}
\end{figure}

\begin{figure}[tb]
        \includegraphics[width = .5\textwidth, trim = 4cm 1.25cm 4cm 1.5cm , clip]{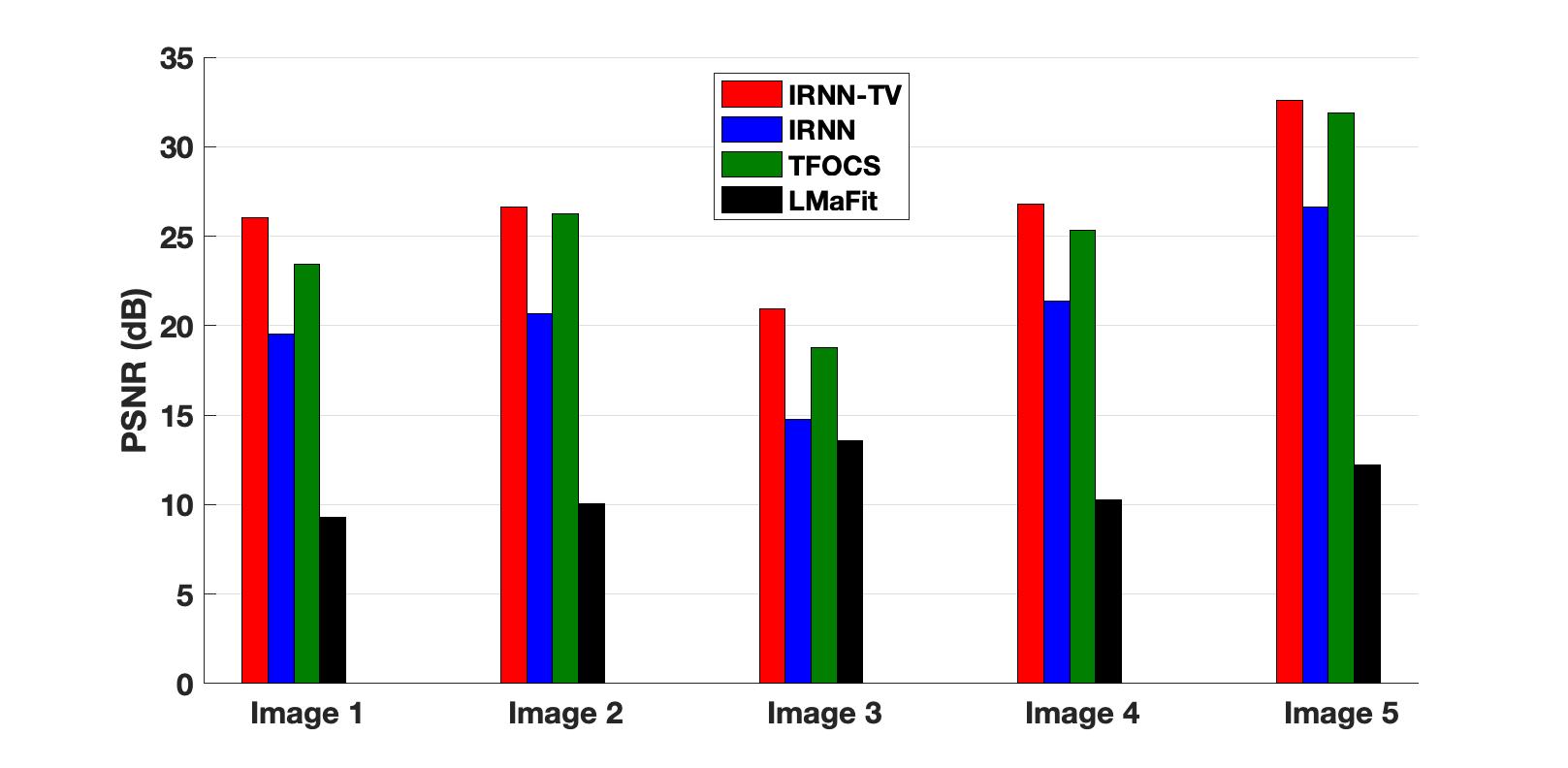}
    \includegraphics[width = 0.5\textwidth, trim = 4cm 1.25cm 4cm 1.5cm , clip]{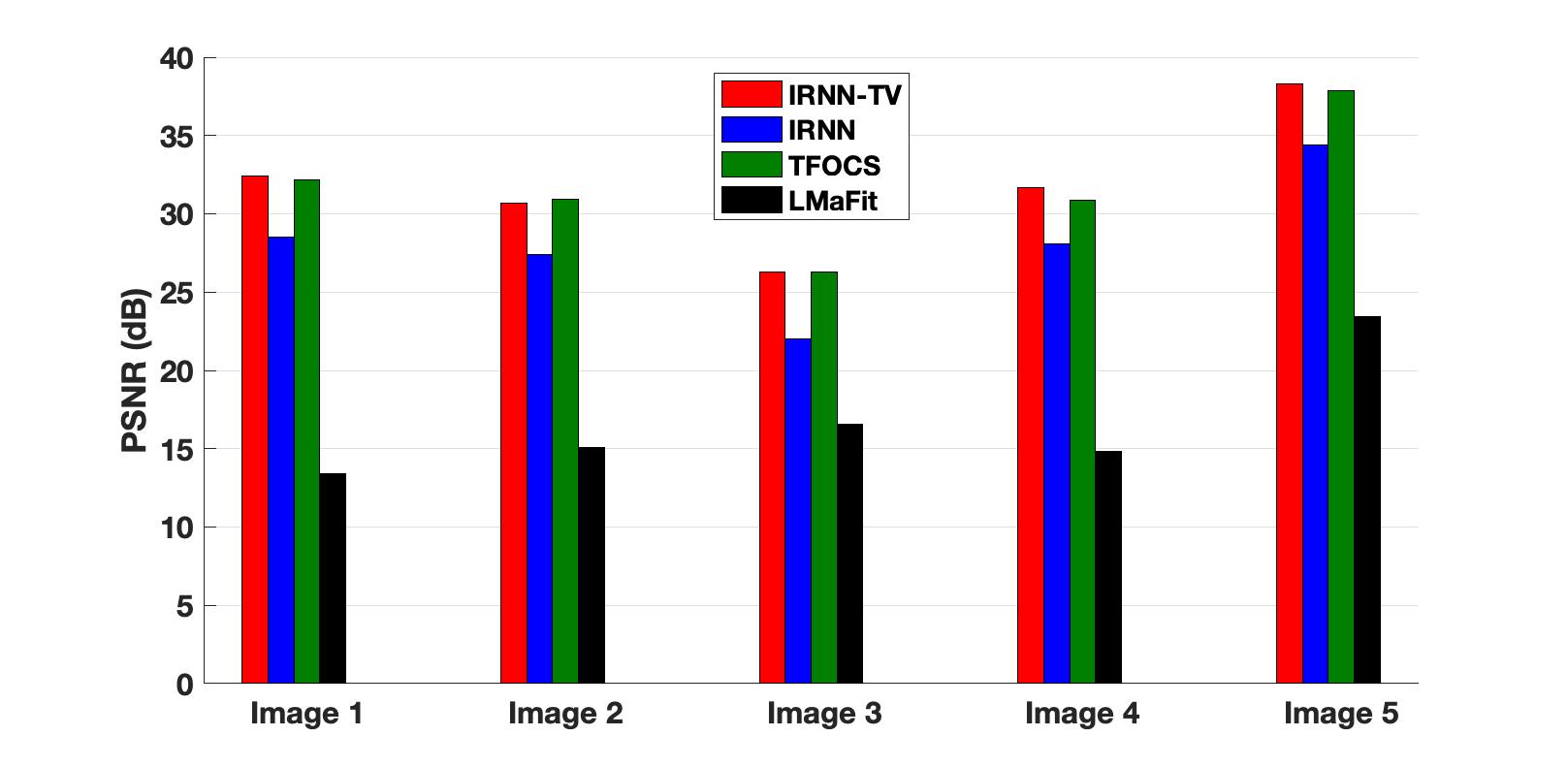}
    \caption{Comparison of the quality of the recovered images when $20\%$ data (in the left) and $50\%$ data (in the right) are observed }
    \label{fig:PSNR_completion}
\end{figure}

\Cref{fig:completion_20,fig:completion_50} present the recovered images by using different techniques.
Since these images, in general, do not have a low-rank structure --- though the singular values decay rather rapidly, recovering images based only on the rank function or one of its associated relaxation techniques such as the nuclear-norm minimization is not enough.
This can be seen in the recovered images by LMaFit and IRNN.
TFOCS which is based on minimizing the total variation norm performs relatively well.
However, it fails sometimes to recover fine features especially when the observed data is small.
\Cref{fig:completion_20_percent_TFOCS_IRNN_TV} illustrates how even with $20\%$ of the original data, our method can recover fine features, even better than TFOCS.
It demonstrates the effectiveness of our method in recovering images using partial observations. It can also be seen in \Cref{fig:PSNR_completion} that presents the PSNR values for each method used in our numerical experiments.

\subsection{Image Completion with Noisy Observations}
In this section, we consider the recovery of an image starting from partially observed contaminated data.
The contamination is performed by adding random noise to the original image such that the PSNR  is 20 dB.
We again observe $20\%$ and $50\%$ data but this time, the contamination is done by adding random noise such that the peak signal to noise ratio is 20 dB.
 We again compare our proposed method against the aforementioned methods.

\begin{figure}[tb]
  \begin{minipage}{0.16\linewidth}
    \begin{adjustbox}{width=\linewidth}
        \includegraphics{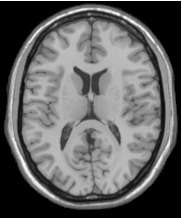}
    \end{adjustbox}
  \end{minipage}
  \begin{minipage}{0.16\linewidth}
    \begin{adjustbox}{width=\linewidth}
        \includegraphics{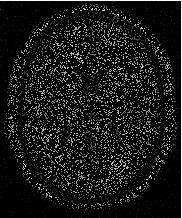}
    \end{adjustbox}
  \end{minipage}
  \begin{minipage}{0.16\linewidth}
    \begin{adjustbox}{width=\linewidth}
        \includegraphics{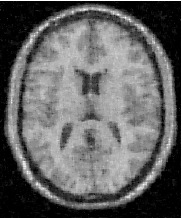}
    \end{adjustbox}
  \end{minipage}
  \begin{minipage}{0.16\linewidth}
    \begin{adjustbox}{width=\linewidth}
        \includegraphics{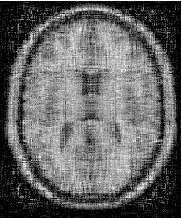}
    \end{adjustbox}
  \end{minipage}
  \begin{minipage}{0.16\linewidth}
    \begin{adjustbox}{width=\linewidth}
        \includegraphics{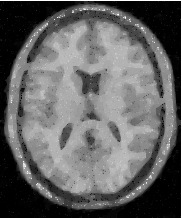}
    \end{adjustbox}
  \end{minipage}
  \begin{minipage}{0.16\linewidth}
    \begin{adjustbox}{width=\linewidth}
        \includegraphics{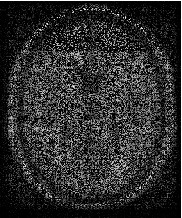}
    \end{adjustbox}
  \end{minipage}

  \begin{minipage}{0.16\linewidth}
    \begin{adjustbox}{width=\linewidth}
        \includegraphics{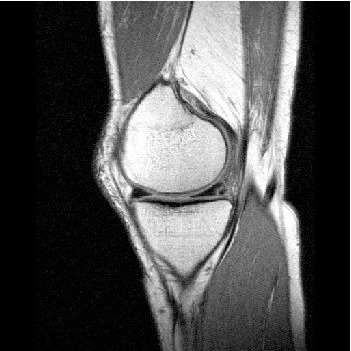}
    \end{adjustbox}
  \end{minipage}
  \begin{minipage}{0.16\linewidth}
    \begin{adjustbox}{width=\linewidth}
        \includegraphics{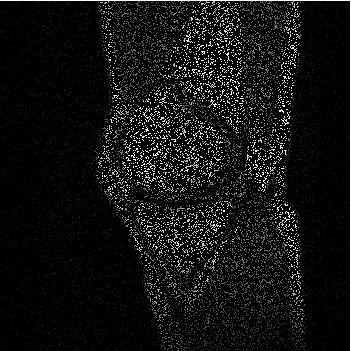}
    \end{adjustbox}
  \end{minipage}
  \begin{minipage}{0.16\linewidth}
    \begin{adjustbox}{width=\linewidth}
        \includegraphics{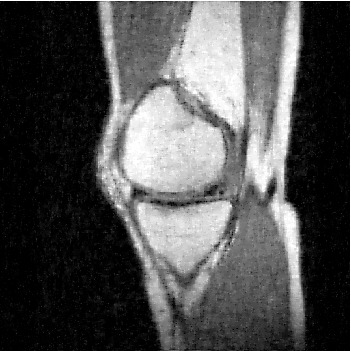}
    \end{adjustbox}
  \end{minipage}
  \begin{minipage}{0.16\linewidth}
    \begin{adjustbox}{width=\linewidth}
        \includegraphics{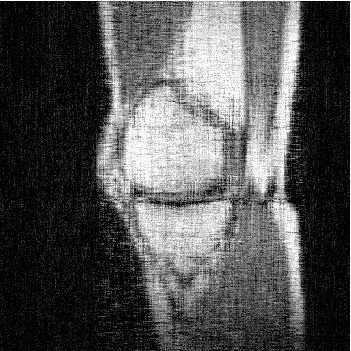}
    \end{adjustbox}
  \end{minipage}
  \begin{minipage}{0.16\linewidth}
    \begin{adjustbox}{width=\linewidth}
        \includegraphics{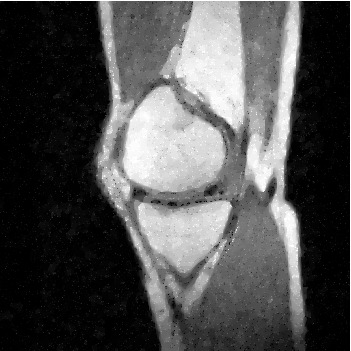}
    \end{adjustbox}
  \end{minipage}
  \begin{minipage}{0.16\linewidth}
    \begin{adjustbox}{width=\linewidth}
        \includegraphics{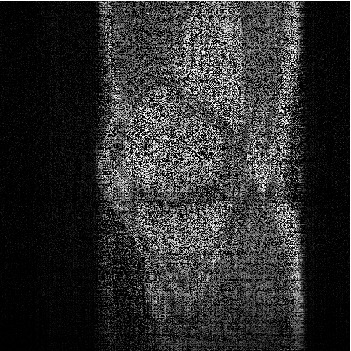}
    \end{adjustbox}
  \end{minipage}

  \begin{minipage}{0.16\linewidth}
    \begin{adjustbox}{width=\linewidth}
        \includegraphics{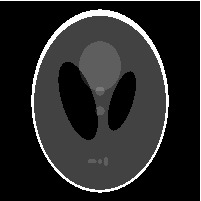}
    \end{adjustbox}
  \end{minipage}
  \begin{minipage}{0.16\linewidth}
    \begin{adjustbox}{width=\linewidth}
        \includegraphics{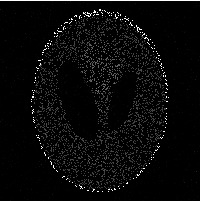}
    \end{adjustbox}
  \end{minipage}
  \begin{minipage}{0.16\linewidth}
    \begin{adjustbox}{width=\linewidth}
        \includegraphics{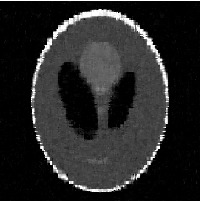}
    \end{adjustbox}
  \end{minipage}
  \begin{minipage}{0.16\linewidth}
    \begin{adjustbox}{width=\linewidth}
        \includegraphics{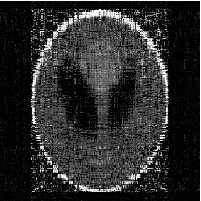}
    \end{adjustbox}
  \end{minipage}
  \begin{minipage}{0.16\linewidth}
    \begin{adjustbox}{width=\linewidth}
        \includegraphics{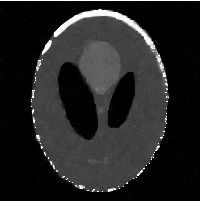}
    \end{adjustbox}
  \end{minipage}
  \begin{minipage}{0.16\linewidth}
    \begin{adjustbox}{width=\linewidth}
        \includegraphics{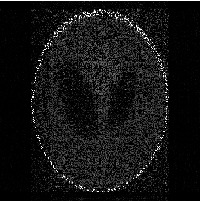}
    \end{adjustbox}
  \end{minipage}

  \begin{minipage}{0.16\linewidth}
    \begin{adjustbox}{width=\linewidth}
        \includegraphics{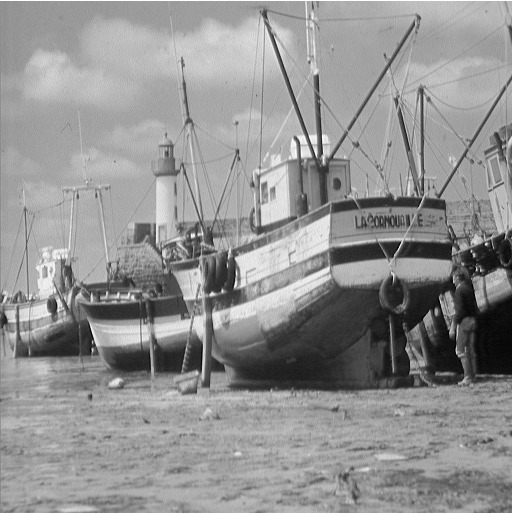}
    \end{adjustbox}
  \end{minipage}
  \begin{minipage}{0.16\linewidth}
    \begin{adjustbox}{width=\linewidth}
        \includegraphics{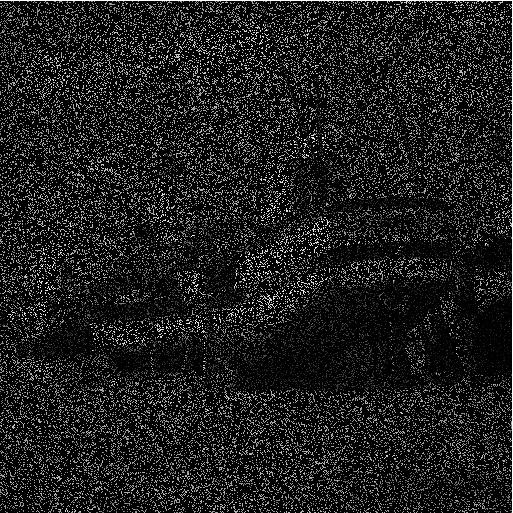}
    \end{adjustbox}
  \end{minipage}
  \begin{minipage}{0.16\linewidth}
    \begin{adjustbox}{width=\linewidth}
        \includegraphics{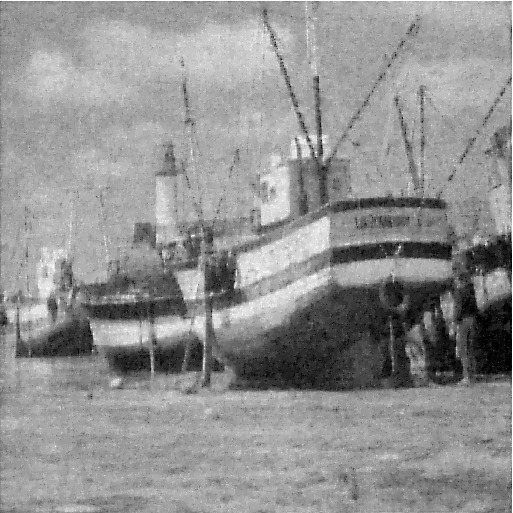}
    \end{adjustbox}
  \end{minipage}
  \begin{minipage}{0.16\linewidth}
    \begin{adjustbox}{width=\linewidth}
        \includegraphics{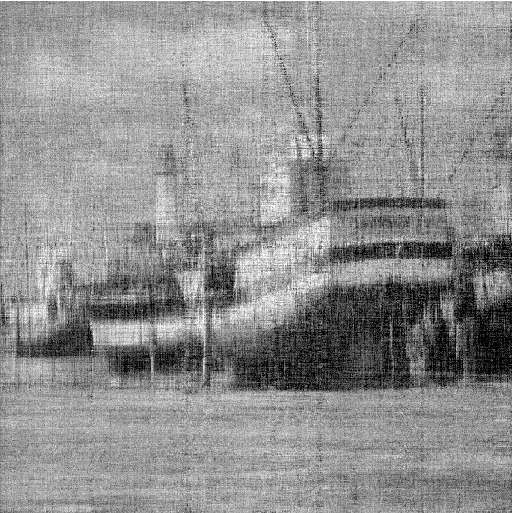}
    \end{adjustbox}
  \end{minipage}
  \begin{minipage}{0.16\linewidth}
    \begin{adjustbox}{width=\linewidth}
        \includegraphics{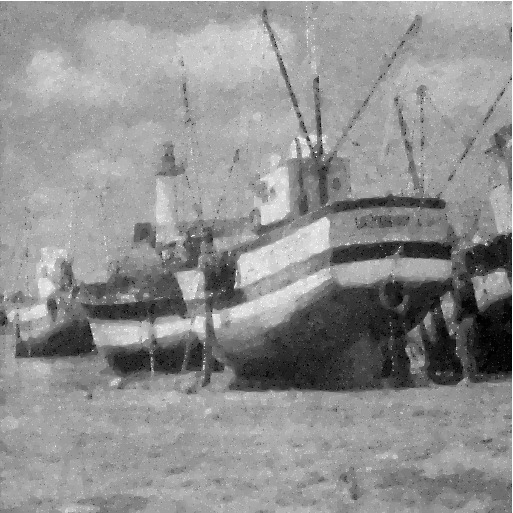}
    \end{adjustbox}
  \end{minipage}
  \begin{minipage}{0.16\linewidth}
    \begin{adjustbox}{width=\linewidth}
        \includegraphics{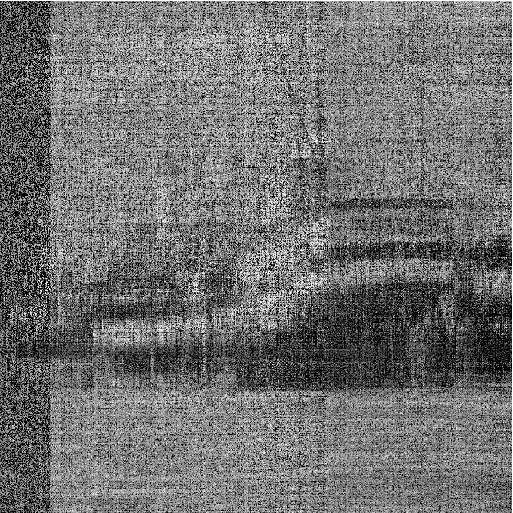}
    \end{adjustbox}
  \end{minipage}

  \begin{minipage}{0.16\linewidth}
    \begin{adjustbox}{width=\linewidth}
        \includegraphics{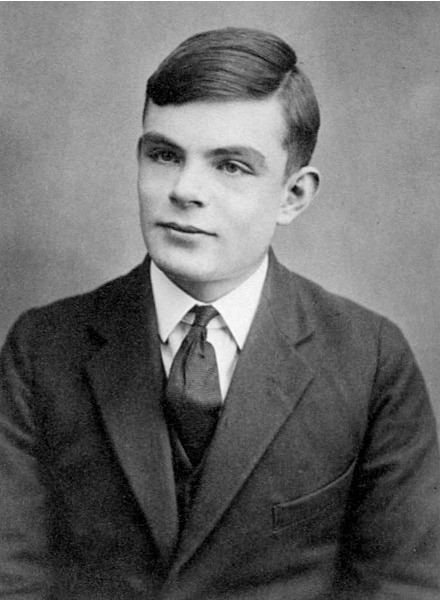}
    \end{adjustbox}
  \end{minipage}
  \begin{minipage}{0.16\linewidth}
    \begin{adjustbox}{width=\linewidth}
        \includegraphics{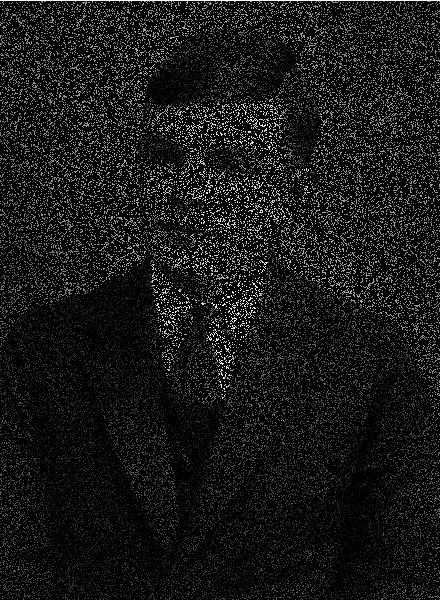}
    \end{adjustbox}
  \end{minipage}
  \begin{minipage}{0.16\linewidth}
    \begin{adjustbox}{width=\linewidth}
        \includegraphics{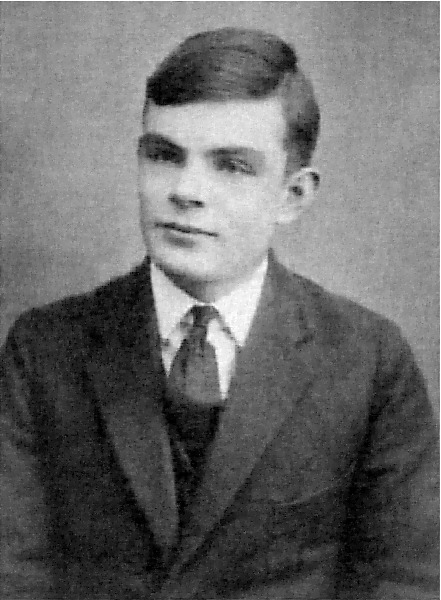}
    \end{adjustbox}
  \end{minipage}
  \begin{minipage}{0.16\linewidth}
    \begin{adjustbox}{width=\linewidth}
        \includegraphics{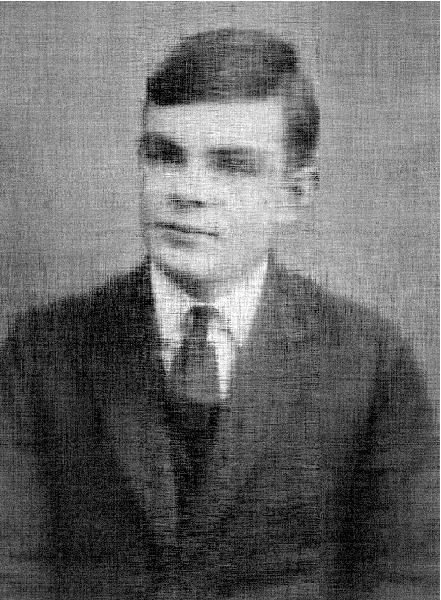}
    \end{adjustbox}
  \end{minipage}
  \begin{minipage}{0.16\linewidth}
    \begin{adjustbox}{width=\linewidth}
        \includegraphics{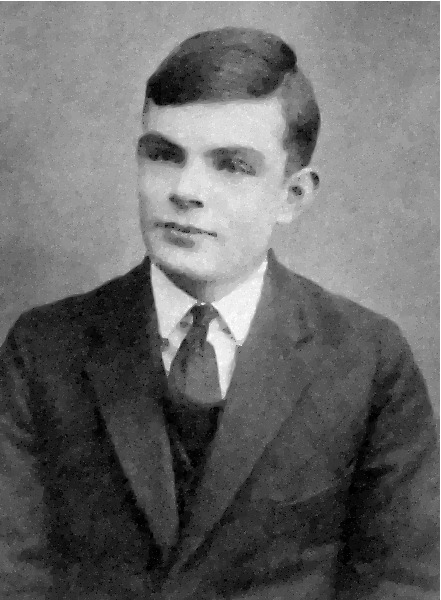}
    \end{adjustbox}
  \end{minipage}
  \begin{minipage}{0.16\linewidth}
    \begin{adjustbox}{width=\linewidth}
        \includegraphics{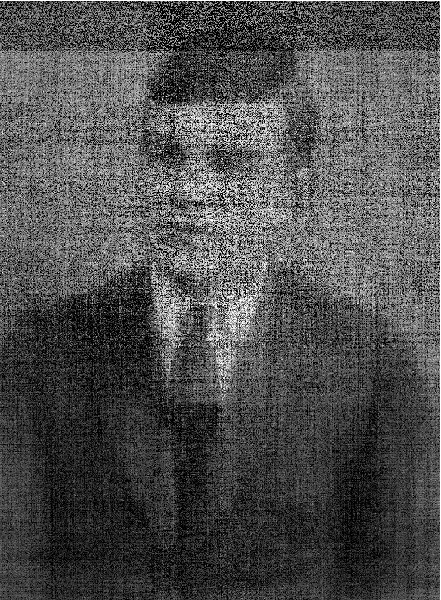}
    \end{adjustbox}
  \end{minipage}
  \caption{Comparison of image recovery by using different techniques when $20\%$ data are observed with some noise (PSNR = 20).
  From the left to the right in each row: original image, observed image, the recovered image by IRNN\_TV, IRNN, TFOCS, and LMaFit, respectively}
  \label{fig:deblurring_20}
\end{figure}

\begin{figure}[tb]
  \begin{minipage}{0.16\linewidth}
    \begin{adjustbox}{width=\linewidth}
        \includegraphics{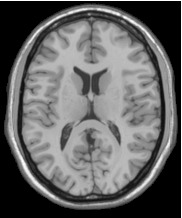}
    \end{adjustbox}
  \end{minipage}
  \begin{minipage}{0.16\linewidth}
    \begin{adjustbox}{width=\linewidth}
        \includegraphics{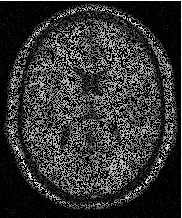}
    \end{adjustbox}
  \end{minipage}
  \begin{minipage}{0.16\linewidth}
    \begin{adjustbox}{width=\linewidth}
        \includegraphics{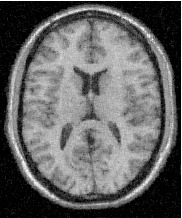}
    \end{adjustbox}
  \end{minipage}
  \begin{minipage}{0.16\linewidth}
    \begin{adjustbox}{width=\linewidth}
        \includegraphics{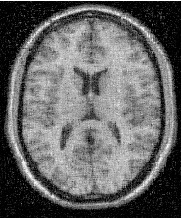}
    \end{adjustbox}
  \end{minipage}
  \begin{minipage}{0.16\linewidth}
    \begin{adjustbox}{width=\linewidth}
        \includegraphics{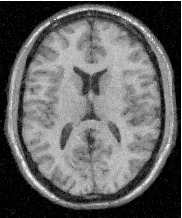}
    \end{adjustbox}
  \end{minipage}
  \begin{minipage}{0.16\linewidth}
    \begin{adjustbox}{width=\linewidth}
        \includegraphics{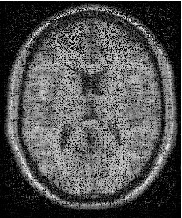}
    \end{adjustbox}
  \end{minipage}

  \begin{minipage}{0.16\linewidth}
    \begin{adjustbox}{width=\linewidth}
        \includegraphics{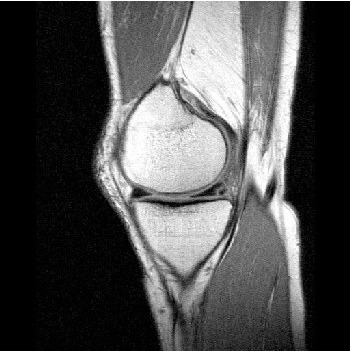}
    \end{adjustbox}
  \end{minipage}
  \begin{minipage}{0.16\linewidth}
    \begin{adjustbox}{width=\linewidth}
        \includegraphics{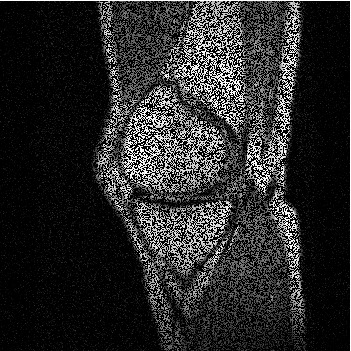}
    \end{adjustbox}
  \end{minipage}
  \begin{minipage}{0.16\linewidth}
    \begin{adjustbox}{width=\linewidth}
        \includegraphics{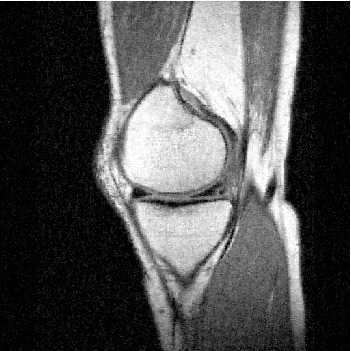}
    \end{adjustbox}
  \end{minipage}
  \begin{minipage}{0.16\linewidth}
    \begin{adjustbox}{width=\linewidth}
        \includegraphics{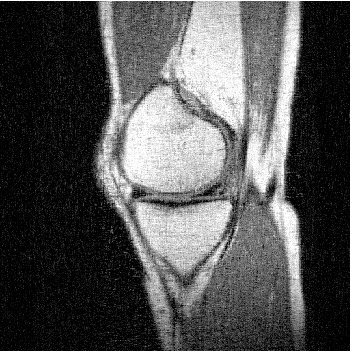}
    \end{adjustbox}
  \end{minipage}
  \begin{minipage}{0.16\linewidth}
    \begin{adjustbox}{width=\linewidth}
        \includegraphics{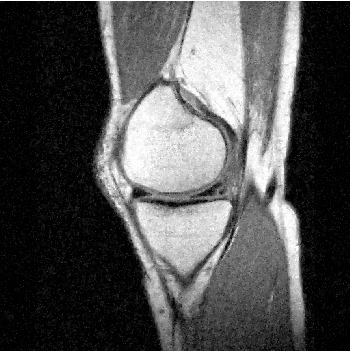}
    \end{adjustbox}
  \end{minipage}
  \begin{minipage}{0.16\linewidth}
    \begin{adjustbox}{width=\linewidth}
        \includegraphics{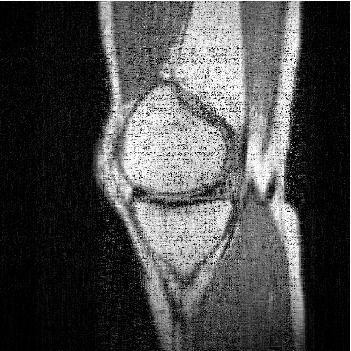}
    \end{adjustbox}
  \end{minipage}

  \begin{minipage}{0.16\linewidth}
    \begin{adjustbox}{width=\linewidth}
        \includegraphics{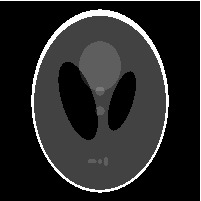}
    \end{adjustbox}
  \end{minipage}
  \begin{minipage}{0.16\linewidth}
    \begin{adjustbox}{width=\linewidth}
        \includegraphics{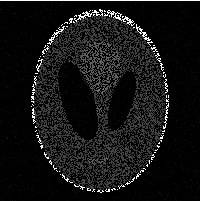}
    \end{adjustbox}
  \end{minipage}
  \begin{minipage}{0.16\linewidth}
    \begin{adjustbox}{width=\linewidth}
        \includegraphics{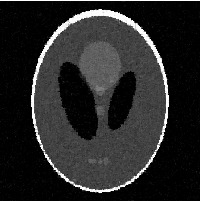}
    \end{adjustbox}
  \end{minipage}
  \begin{minipage}{0.16\linewidth}
    \begin{adjustbox}{width=\linewidth}
        \includegraphics{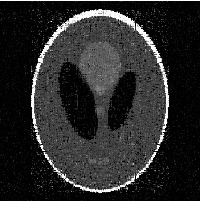}
    \end{adjustbox}
  \end{minipage}
  \begin{minipage}{0.16\linewidth}
    \begin{adjustbox}{width=\linewidth}
        \includegraphics{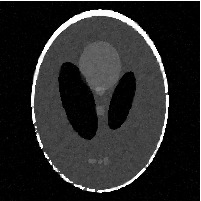}
    \end{adjustbox}
  \end{minipage}
  \begin{minipage}{0.16\linewidth}
    \begin{adjustbox}{width=\linewidth}
        \includegraphics{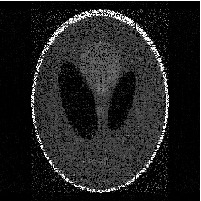}
    \end{adjustbox}
  \end{minipage}

  \begin{minipage}{0.16\linewidth}
    \begin{adjustbox}{width=\linewidth}
        \includegraphics{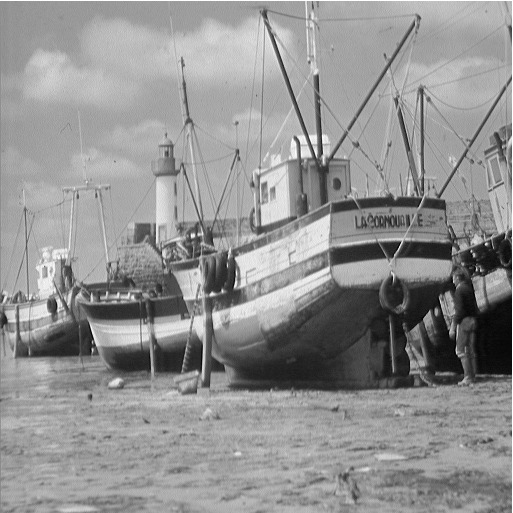}
    \end{adjustbox}
  \end{minipage}
  \begin{minipage}{0.16\linewidth}
    \begin{adjustbox}{width=\linewidth}
        \includegraphics{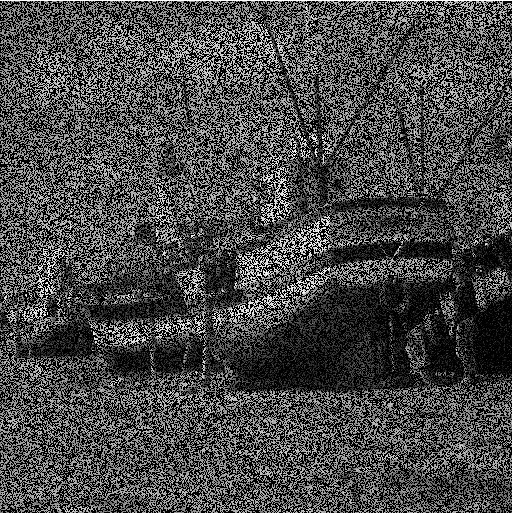}
    \end{adjustbox}
  \end{minipage}
  \begin{minipage}{0.16\linewidth}
    \begin{adjustbox}{width=\linewidth}
        \includegraphics{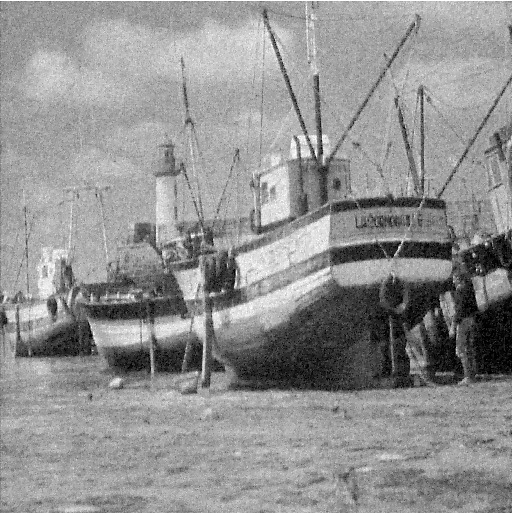}
    \end{adjustbox}
  \end{minipage}
  \begin{minipage}{0.16\linewidth}
    \begin{adjustbox}{width=\linewidth}
        \includegraphics{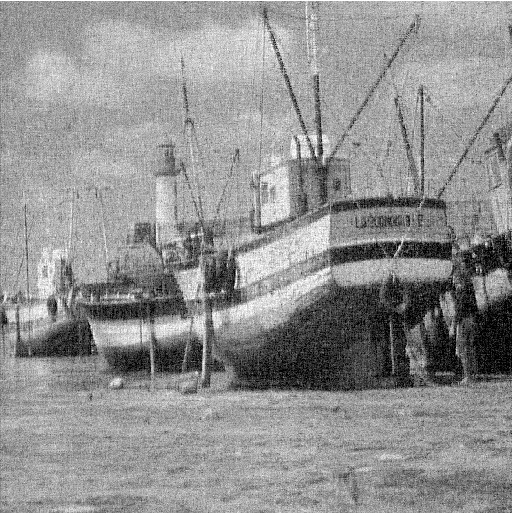}
    \end{adjustbox}
  \end{minipage}
  \begin{minipage}{0.16\linewidth}
    \begin{adjustbox}{width=\linewidth}
        \includegraphics{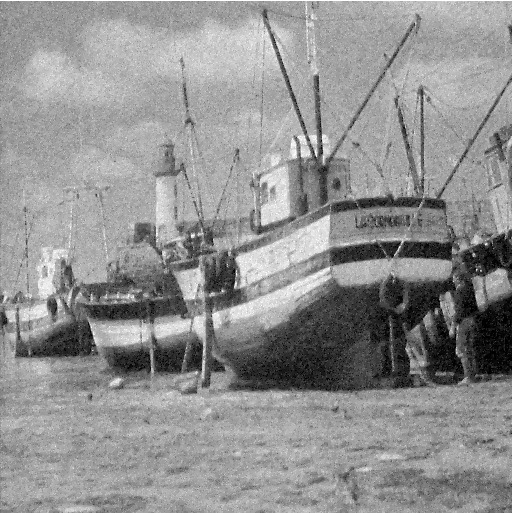}
    \end{adjustbox}
  \end{minipage}
  \begin{minipage}{0.16\linewidth}
    \begin{adjustbox}{width=\linewidth}
        \includegraphics{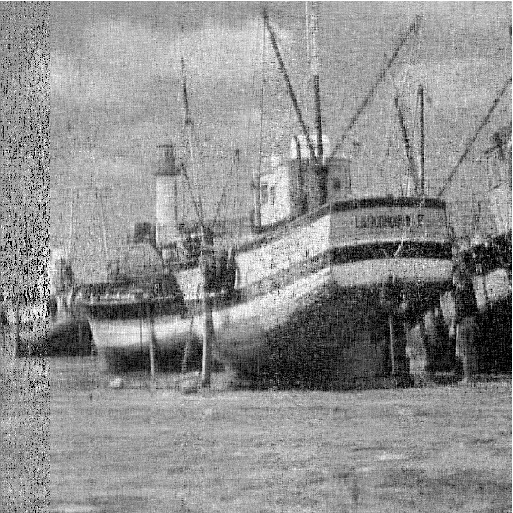}
    \end{adjustbox}
  \end{minipage}

  \begin{minipage}{0.16\linewidth}
    \begin{adjustbox}{width=\linewidth}
        \includegraphics{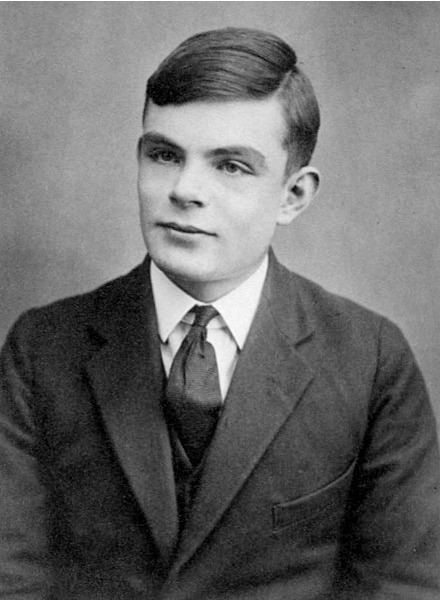}
    \end{adjustbox}
  \end{minipage}
  \begin{minipage}{0.16\linewidth}
    \begin{adjustbox}{width=\linewidth}
        \includegraphics{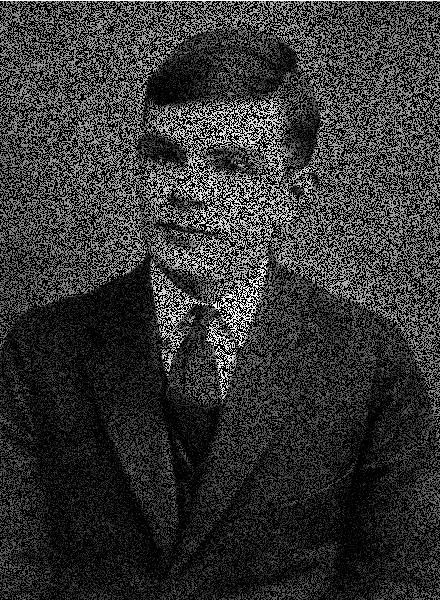}
    \end{adjustbox}
  \end{minipage}
  \begin{minipage}{0.16\linewidth}
    \begin{adjustbox}{width=\linewidth}
        \includegraphics{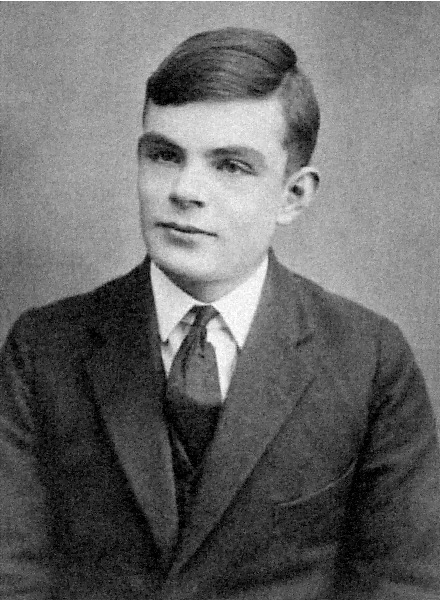}
    \end{adjustbox}
  \end{minipage}
  \begin{minipage}{0.16\linewidth}
    \begin{adjustbox}{width=\linewidth}
        \includegraphics{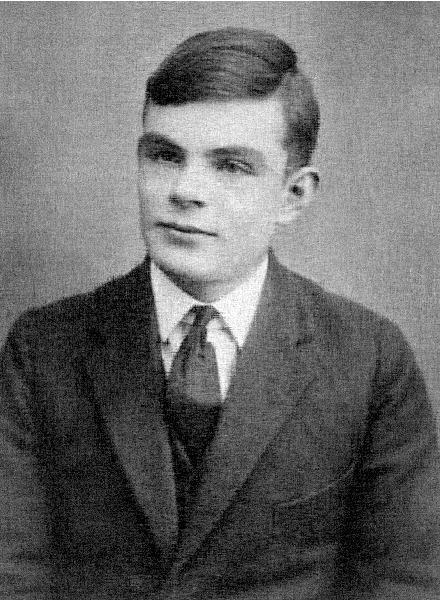}
    \end{adjustbox}
  \end{minipage}
  \begin{minipage}{0.16\linewidth}
    \begin{adjustbox}{width=\linewidth}
        \includegraphics{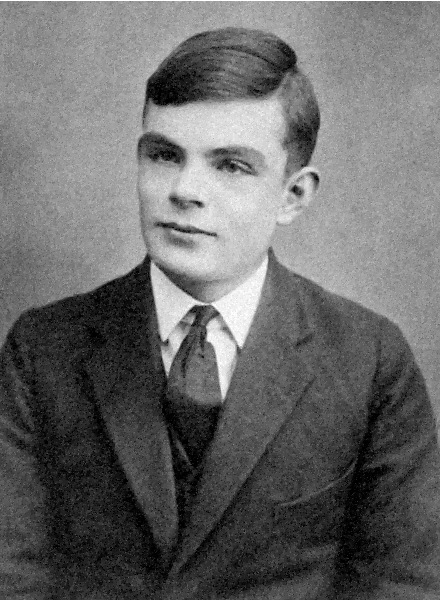}
    \end{adjustbox}
  \end{minipage}
  \begin{minipage}{0.16\linewidth}
    \begin{adjustbox}{width=\linewidth}
        \includegraphics{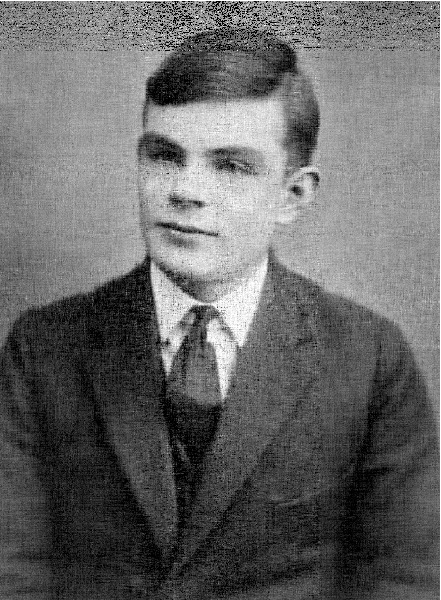}
    \end{adjustbox}
  \end{minipage}
  \caption{Comparison of image recovery by using different techniques when $50\%$ data are observed with some noise (PSNR = 20).
  From the left to the right in each row: original image, observed image, and the recovered image by IRNN\_TV, IRNN, TFOCS, and LMaFit, respectively}
  \label{fig:deblurring_50}
\end{figure}

\begin{figure}[tb]
\centering
        \includegraphics[width = .65\textwidth, trim = 4cm 1.25cm 4cm 1.5cm , clip]{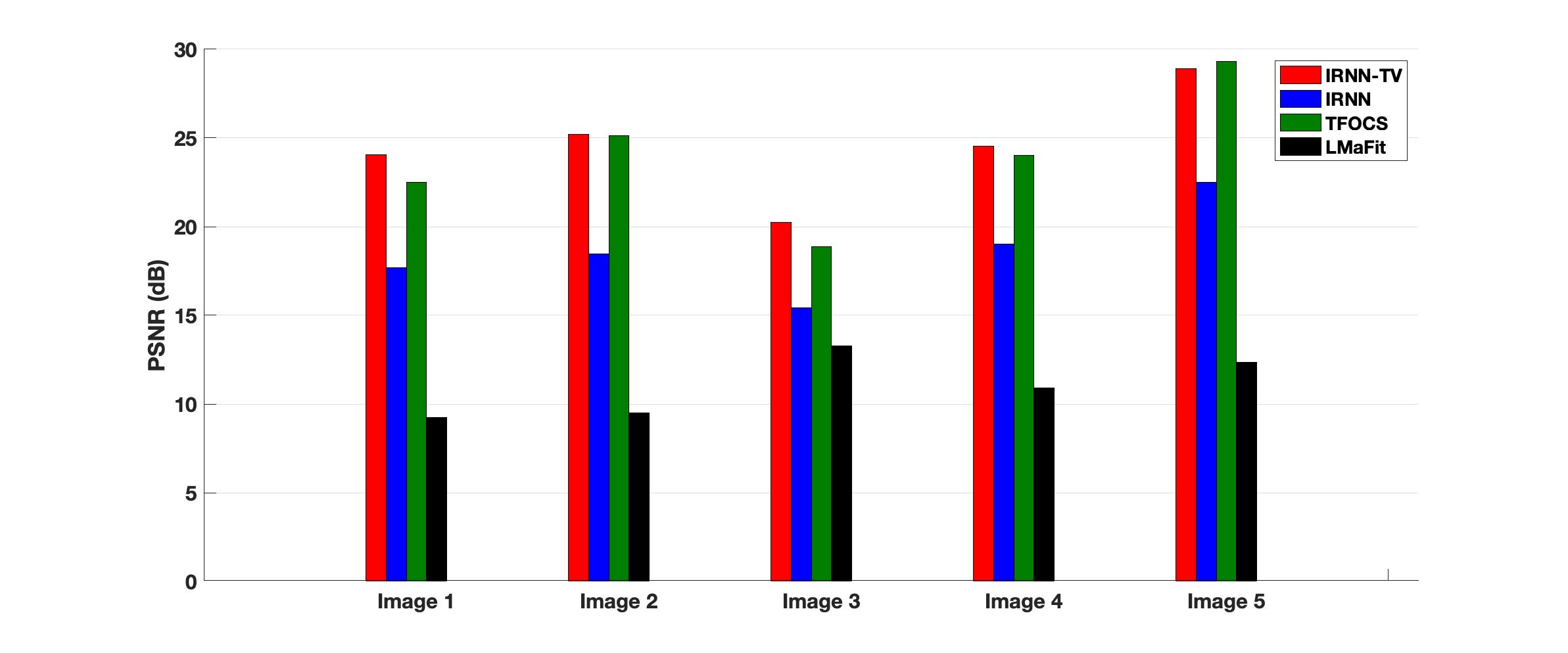}
        
    \includegraphics[width = 0.65\textwidth,trim = 4cm 1.25cm 4cm 1.5cm , clip]{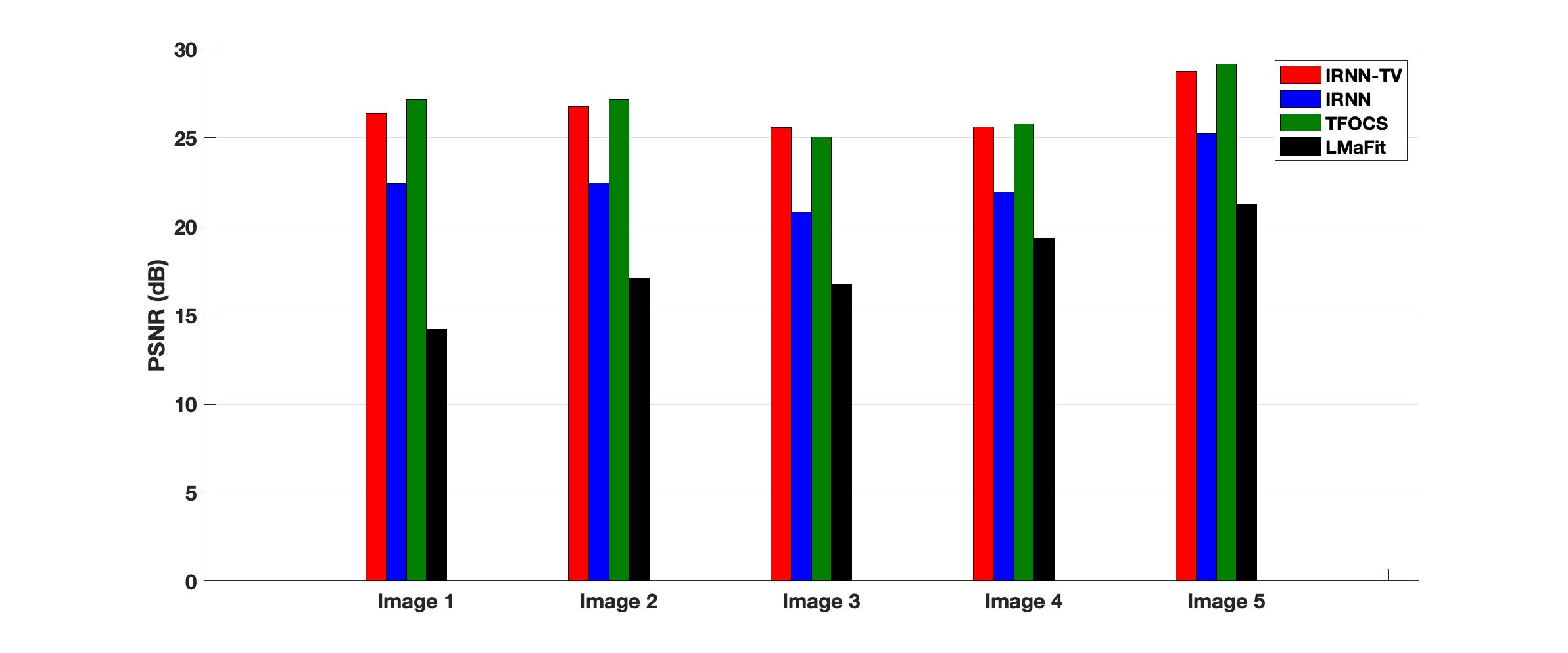}
    \caption{Comparison of the quality of the recovered images when contaminated $20\%$ data (in the top) and $50\%$ data (in the bottom) are observed  }
    \label{fig:PSNR_deblurring}
\end{figure}

\Cref{fig:deblurring_20,fig:deblurring_50} present the noisy observations and recovered images by using different techniques.
Here again, our method and TFOCS show their effectiveness in recovering images.
The PSNR values for each image recovered in the image set by using different techniques are presented in~\Cref{fig:PSNR_deblurring}.

\subsection{Comments on the Implementation for Large-Scale Problems}
The proposed methodology (\Cref{alg:LRTV_IterativeAlgo}) requires the computation of the SVD factorization of subsequent matrices.
This is typically the most computationally expensive step. 
However, our algorithm which is mainly based on the singular values shrinkage-like operator, requires primarily the computation of rank $k$ truncated SVD decomposition for some $k \geq 0$. 

\Cref{tab:rank_completion_20,tab:rank_deblurring_20} present the rank of the recovered images by using the different methods considered in our numerical experiments. First, note that IRNN recovers images with lower ranks than the ones recovered by the proposed method (IRNN\_TV) but the quality of recovered images by IRNN is much poor if compared to IRNN\_TV as reported in the previous two subsections. Therein, we have noticed that IRNN\_TV and TFOCS recover images from partial observations much better than the other two considered methods with IRNN\_TV often being slightly superior.  However, it is also worthwhile to note that the numerical rank of the images recovered using IRNN\_TV is much lower as compared to TFOCS, see again \Cref{tab:rank_completion_20,tab:rank_deblurring_20}. Hence, during the iteration of our proposed \Cref{alg:LRTV_IterativeAlgo}, we need to only keep the solution in a low-rank form, thus requiring lesser storage and reducing computation cost. Moreover, the low-rank factor of the solution at each iteration comes at no additional cost as we employ the singular values thresholding operator at each iteration. Additionally, as discussed in \Cref{rem:rankdec}: for some surrogate functions $g(x)$ of the rank functions, the sequence generated $\bX^k$ by our method will be of non-increasing rank; therefore, an upper bound of $k$, relatively tight, is known a priori. This allows to perform the SVD very efficiently. This can be done, for example, by exploiting iterative and randomized SVD solvers to tackle large scale problems, thus further reducing computation cost.

\begin{table}[tb]
\centering
\makebox[0pt][c]{\parbox{1.0\textwidth}{%
    \begin{minipage}[b]{0.49\hsize}\centering
    {\footnotesize
      \caption{
    Comparison of numerical rank of recovered matrices (images) in \Cref{fig:completion_20}
   % Ordering of the images corresponds to the one in \Cref{fig:completion_20} from top to bottom.
  }
  \label{tab:rank_completion_20}
  \begin{tabular}{|c|c|c|c|c|} \hline
    Image &  IRNN-TV & IRNN &  TFOCS & LMaFit \\
    \hline
     1   & 47  &  31  &  181  &  50 \\ \hline
  	 2   & 98  &  57  &  350  &  52 \\ \hline
  	 3   & 43  &  24  &  195  &  50 \\ \hline
  	 4   & 205 &  95  &  512  &  56 \\ \hline
  	 5   & 166 &  93  &  440  &  58  \\ \hline
  \end{tabular}

  }
    \end{minipage}
    \hfill
    \begin{minipage}[b]{0.49\hsize}\centering
    {\footnotesize
      \caption{
    Comparison of numerical rank of recovered matrices (images) in \Cref{fig:deblurring_20}
   % Ordering of the images corresponds to the one in \Cref{fig:deblurring_20} from top to bottom.
  }
  \label{tab:rank_deblurring_20}
  }
  \begin{tabular}{|c|c|c|c|c|} \hline
    Image &  IRNN-TV & IRNN &  TFOCS & LMaFit \\
    \hline
     1   & 64  & 33  &  181  & 56 \\ \hline
  	 2   & 155 & 61  &  350  & 60 \\ \hline
  	 3   & 63  & 31  &  200  & 54 \\ \hline
  	 4   & 276 & 95  &  512  & 70 \\ \hline
  	 5   & 260 & 93  &  440  & 70  \\ \hline
  \end{tabular}
    \end{minipage}
}}
\end{table}

\section{Conclusions}\label{sec:conclusions}
In this paper, we have studied the image recovery problem. 
For this, we have proposed an optimization problem using a combination of low-rank and total variation regularizers; hence, it is expected to capture both, spatially local and global features, of the image better than if only one of the regularizers is considered.
Furthermore, we have proposed an iterative scheme to solve such an optimization problem that essentially requires to apply weighted singular value thresholding at each iteration. 
And the convergence of the iterative scheme is guaranteed. 
Finally, we have demonstrated that the proposed method outperforms when compared to state-of-the-art methods.  
In our future work, we seek to study a similar problem with applications to 3-dimensional objects and denoising video surveillance while incorporating tensor techniques. 

\clearpage
\bibliographystyle{splncs04}
\bibliography{egbib}

\begin{thebibliography}{10}
\providecommand{\url}[1]{\texttt{#1}}
\providecommand{\urlprefix}{URL }
\providecommand{\doi}[1]{https://doi.org/#1}

\bibitem{TFOCS}
Becker, S.R., Cand{\`e}s, E.J., Grant, M.C.: Templates for convex cone problems
  with applications to sparse signal recovery. Mathematical {P}rogramming
  {C}omputation  \textbf{3}(3), ~165 (2011)

\bibitem{buchanan2005damped}
Buchanan, A.M., Fitzgibbon, A.W.: Damped {N}ewton algorithms for matrix
  factorization with missing data. In: IEEE Computer Society Conference on
  Computer Vision and Pattern Recognition. vol.~2, pp. 316--322. IEEE (2005)

\bibitem{cai2010singular}
Cai, J.F., Cand{\`e}s, E.J., Shen, Z.: A singular value thresholding algorithm
  for matrix completion. SIAM Journal on Optimization  \textbf{20}(4),
  1956--1982 (2010)

\bibitem{candes2009exact}
Cand{\`e}s, E.J., Recht, B.: Exact matrix completion via convex optimization.
  Foundations of Computational Mathematics  \textbf{9}(6), ~717 (2009)

\bibitem{candes2006robust}
Cand{\`e}s, E.J., Romberg, J., Tao, T.: Robust uncertainty principles: Exact
  signal reconstruction from highly incomplete frequency information. IEEE
  Transactions on Information Theory  \textbf{52}(2),  489--509 (2006)

\bibitem{chen2013reduced}
Chen, K., Dong, H., Chan, K.S.: Reduced rank regression via adaptive nuclear
  norm penalization. Biometrika  \textbf{100}(4),  901--920 (2013)

\bibitem{donoho2006compressed}
Donoho, D.L.: Compressed sensing. IEEE Transactions on Information Theory
  \textbf{52}(4),  1289--1306 (2006)

\bibitem{eriksson2010efficient}
Eriksson, A., Van Den~Hengel, A.: Efficient computation of robust low-rank
  matrix approximations in the presence of missing data using the $l_1$ norm.
  In: IEEE Computer Society Conference on Computer Vision and Pattern
  Recognition. pp. 771--778 (2010)

\bibitem{frank1993statistical}
Frank, L.E., Friedman, J.H.: A statistical view of some chemometrics regression
  tools. Technometrics  \textbf{35}(2),  109--135 (1993)

\bibitem{friedman2012fast}
Friedman, J.H.: Fast sparse regression and classification. International
  Journal of Forecasting  \textbf{28}(3),  722--738 (2012)

\bibitem{gao2011feasible}
Gao, C., Wang, N., Yu, Q., Zhang, Z.: A feasible nonconvex relaxation approach
  to feature selection. In: AAAI Conference on Artificial Intelligence (2011)

\bibitem{geman1995nonlinear}
Geman, D., Yang, C.: Nonlinear image recovery with half-quadratic
  regularization. IEEE Transactions on Image Processing  \textbf{4}(7),
  932--946 (1995)

\bibitem{gu2017weighted}
Gu, S., Xie, Q., Meng, D., Zuo, W., Feng, X., Zhang, L.: Weighted nuclear norm
  minimization and its applications to low level vision. International Journal
  of Computer Vision  \textbf{121}(2),  183--208 (2017)

\bibitem{gu2014weighted}
Gu, S., Zhang, L., Zuo, W., Feng, X.: Weighted nuclear norm minimization with
  application to image denoising. In: Proceedings of the IEEE Conference on
  Computer Vision and Pattern Recognition. pp. 2862--2869 (2014)

\bibitem{ke2005robust}
Ke, Q., Kanade, T.: Robust l/sub 1/norm factorization in the presence of
  outliers and missing data by alternative convex programming. In: IEEE
  Computer Society Conference on Computer Vision and Pattern Recognition.
  vol.~1, pp. 739--746 (2005)

\bibitem{lu2015nonconvex}
Lu, C., Tang, J., Yan, S., Lin, Z.: Nonconvex nonsmooth low rank minimization
  via iteratively reweighted nuclear norm. IEEE Transactions on Image
  Processing  \textbf{25}(2),  829--839 (2015)

\bibitem{mu2011accelerated}
Mu, Y., Dong, J., Yuan, X., Yan, S.: Accelerated low-rank visual recovery by
  random projection. In: IEEE Computer Society Conference on Computer Vision
  and Pattern Recognition. pp. 2609--2616. IEEE (2011)

\bibitem{recht2010guaranteed}
Recht, B., Fazel, M., Parrilo, P.A.: Guaranteed minimum-rank solutions of
  linear matrix equations via nuclear norm minimization. SIAM Review
  \textbf{52}(3),  471--501 (2010)

\bibitem{rockafellar1970convex}
Rockafellar, R.T.: Convex {A}nalysis. No.~28, Princeton University Press (1970)

\bibitem{srebro2003weighted}
Srebro, N., Jaakkola, T.: Weighted low-rank approximations. In: Proceedings of
  the 20th International Conference on Machine Learning. pp. 720--727 (2003)

\bibitem{srebro2010collaborative}
Srebro, N., Salakhutdinov, R.R.: Collaborative filtering in a non-uniform
  world: Learning with the weighted trace norm. In: Advances in Neural
  Information Processing Systems. pp. 2056--2064 (2010)

\bibitem{trzasko2008highly}
Trzasko, J., Manduca, A.: Highly undersampled magnetic resonance image
  reconstruction via homotopic $l_0$-minimization. IEEE Transactions on Medical
  imaging  \textbf{28}(1),  106--121 (2008)

\bibitem{LMaFit}
Wen, Z., Yin, W., Zhang, Y.: Solving a low-rank factorization model for matrix
  completion by a nonlinear successive over-relaxation algorithm. Mathematical
  {P}rogramming {C}omputation  \textbf{4}(4),  333--361 (2012)

\bibitem{wright2009robust}
Wright, J., Ganesh, A., Rao, S., Peng, Y., Ma, Y.: Robust principal component
  analysis: Exact recovery of corrupted low-rank matrices via convex
  optimization. In: Advances in Neural Information Processing Systems. pp.
  2080--2088 (2009)

\bibitem{zhang2010nearly}
Zhang, C.H., et~al.: Nearly unbiased variable selection under minimax concave
  penalty. The Annals of Statistics  \textbf{38}(2),  894--942 (2010)

\bibitem{zhang2010analysis}
Zhang, T.: Analysis of multi-stage convex relaxation for sparse regularization.
  Journal of Machine Learning Research  \textbf{11},  1081--1107 (2010)

\end{thebibliography}

 \end{document}